\documentclass[11pt]{article}
\usepackage[preprint]{acl}
\usepackage{times,latexsym}
\usepackage[T1]{fontenc}
\usepackage[utf8]{inputenc}

\usepackage{amsmath,amsfonts,bm}

\def\eqref#1{equation~\ref{#1}}

\def\1{\bm{1}}

\DeclareMathAlphabet{\mathsfit}{\encodingdefault}{\sfdefault}{m}{sl}
\SetMathAlphabet{\mathsfit}{bold}{\encodingdefault}{\sfdefault}{bx}{n}

\usepackage{amsmath,amssymb,mathtools}
\usepackage{float}
\usepackage{graphicx}
\usepackage{multirow}
\usepackage{algorithm}
\usepackage{algpseudocode}
\usepackage{microtype}
\usepackage{array}
\usepackage{longtable}
\usepackage{booktabs}

\newcommand{\sg}{\operatorname{sg}}
\newcommand{\method}{DualOPSD}
\newcommand{\methodplain}{DualOPSD w/o Residual}

\title{DualOPSD: Closing the Feedback Loop in On-Policy Self-Distillation}

\author{Yutong Chen\textsuperscript{1} \quad Guangfu Guo\textsuperscript{1} \quad
Zhichao Xu\textsuperscript{2} \quad Kunpeng Liu\textsuperscript{1}\thanks{Corresponding author.} \\
\textsuperscript{1}Department of Computer Science, Clemson University\\
\textsuperscript{2}Kahlert School of Computing, University of Utah\\
\texttt{\{yutong2, gguo, kunpenl\}@clemson.edu}\\
\texttt{zhichao.xu@utah.edu}
}

\begin{document}
\maketitle
\begin{abstract}
On-policy self-distillation (OPSD) combines student generated trajectories with dense supervision from a privileged teacher of the same model. Its pointwise clipping protects the student from locally incompatible teacher directions, but the fixed teacher receives no feedback about the supervision the student rejects, which limits effective learning from the teacher. We introduce \method{}, which alternates updates in a coupled teacher--student system. The student retains OPSD's clipped forward KL objective. The privileged teacher adapts to the updated student after each student step. A residual target based on the frozen initial model preserves the effect of privileged context during teacher adaptation. 
Across Qwen3-8B/4B/1.7B and three math benchmarks, \method{} improves avg@12 accuracy over OPSD by 9.83 percentage points. Further diagnostic analyses show lower bidirectional KL, less persistent mismatch, and retained sensitivity to privileged reference. Together, these results demonstrate that \method{} can be a more effective form of privileged self-distillation.

Code is available at: \url{https://github.com/on1262/DualOPSD}.
\end{abstract}

\section{Introduction}

On-policy self-distillation (OPSD) combines on-policy training with dense teacher supervision~\citep{zhao2026opsd,agarwal2024onpolicy,gu2023minillm}. A student generates a trajectory, and the same model conditioned on a verified reference solution provides a full next token distribution at each prefix generated by the student. Unlike supervised finetuning, OPSD trains the student on states induced by its current policy~\citep{nie2026states}. Compared with reinforcement learning with verifiable rewards, OPSD provides denser supervision because every generated token receives feedback~\citep{shao2024deepseekmath,guo2025deepseekr1}.

OPSD uses pointwise clipping to suppress large KL contributions of style tokens at the vocabulary level. This protects the student from being misled by teacher styles~\citep{zhao2026opsd}. However, once a contribution is clipped, the teacher receives no feedback about the rejected supervision. As training changes the student distribution and the prefixes it visits, the fixed teacher can continue to produce incompatible directions, which eventually leads to performance degradation. More broadly, analyses of on-policy distillation find that effective transfer requires the teacher and student to share compatible thinking patterns~\citep{li2026rethinkingopd}. Therefore, the teacher-student mismatch limits student learning in on-policy self-distillation.

We examine this mismatch in Section~\ref{sec:fixed-diagnostic}. On Qwen3-4B, $D_{\mathrm{KL}}(p_T\|p_S)$, which measures the divergence between two distributions, grows by roughly an order of magnitude within 100 training steps, and dominant clipped teacher directions recur on the same problems.
These observations suggest that clipping mechanism repeatedly suppresses conflicts without addressing the root cause: which is the conflict between a student model being updated and a static teacher through the course of the training.

We therefore ask how we can adapt the teacher to the evolving student during OPSD, while still maintaining the privileged signal for meaningful supervision. Updating the teacher toward the student provides a feedback path, but may also weaken its sensitivity to privileged context, eventually lower the performance ceiling.

\begin{figure*}[t]
    \centering
    \includegraphics[width=\textwidth]{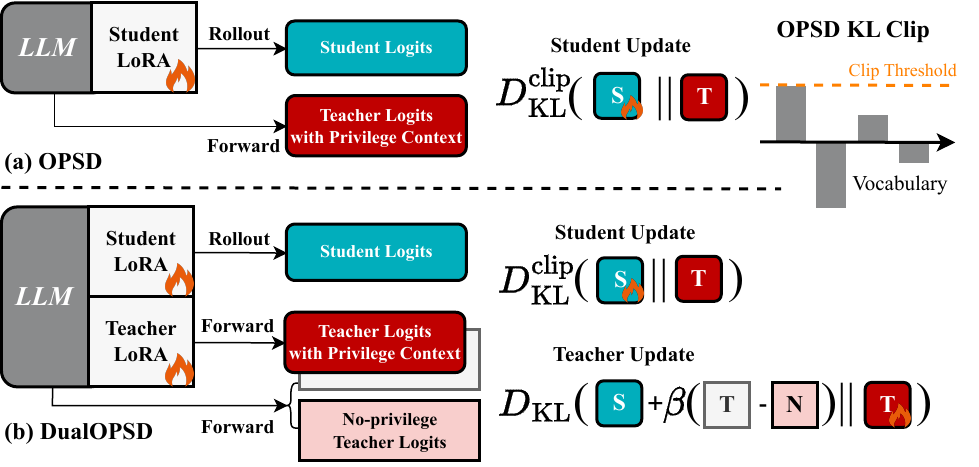}
    \caption{\textbf{Overview of \method{}.} \textbf{(a)} OPSD updates only the student by minimizing a pointwise clipped forward KL divergence to a fixed privileged teacher. \textbf{(b)} In \method{}, the student retains OPSD's clipped forward KL update. The teacher adapts toward the updated student while adding a scaled residual, which is used to preserve the privileged effect during adaptation. Flames denote where gradients are propagated.}
    \label{fig:overview}
\end{figure*}

To this end, we propose \method{}, an asymmetric alternating upstate strategy, aiming to adapt the teacher to the student while preserving the effect of information available only to the teacher. Figure~\ref{fig:overview} shows an illustration of \method{}. The student retains OPSD's clipped forward KL objective. After the student update, the teacher is optimized on the same trajectory using the updated student distribution. To preserve the privileged effect during adaptation, we add the frozen initial model's log-probability residual to the updated student's log probabilities and normalize the result to form the teacher target. The teacher can thus respond to changes in the student while remaining anchored to the initial effect of privileged context.

Across three popular math benchmarks, AIME 2024, AIME 2025, and HMMT, \method{} consistently improves over OPSD at all three model scales. Averaged across the three benchmarks, the gains are about 1.5 percentage points at 1.7B, 12 at 4B, and 16 at 8B, or nearly 10 points across scales. The gains persist under matched compute. Overall response length efficiency is about twice that of OPSD. 
On Qwen3-4B at step 100, the forward KL between teacher and student falls to 37.4\% of the OPSD value, and the recurrence of clipped forward KL directions falls by more than threefold.
In our ablation studies, the residual target nearly doubles sensitivity to privileged information.
Overall, these results show that \method{} improves accuracy and efficiency while producing more stable teacher--student alignment.

Our contributions are:
\begin{itemize}
    \item We diagnose growing teacher--student mismatch and persistent clipped teacher directions under OPSD, identifying feedback that the student update suppresses but does not return to the teacher.
    \item We introduce asymmetric alternating updates that adapt the teacher to the updated student while preserving the initial effect of privileged context through a frozen residual target.
    \item We evaluate accuracy, distributional alignment, privileged sensitivity, generation behavior, and training cost across three Qwen3 scales, finding substantial gains at 4B/8B and a smaller improvement over OPSD at 1.7B.
\end{itemize}

\begin{figure*}[t]
    \centering
    \includegraphics[width=\textwidth]{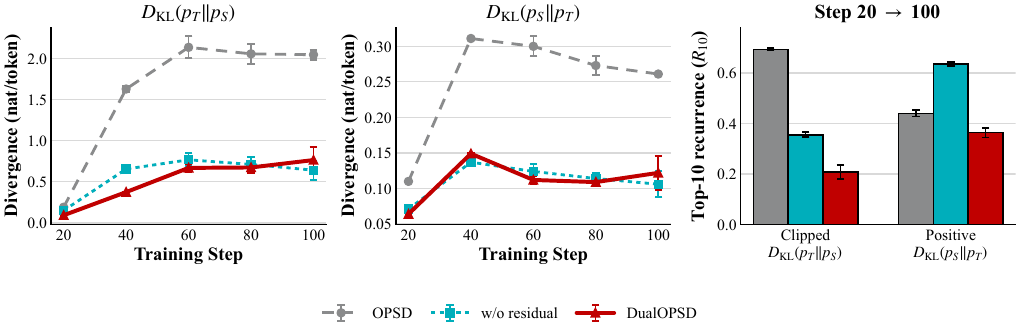}
    \caption{Distributional mismatch and temporal recurrence on Qwen3-4B. OPSD has a larger distributional gap and more recurrent clipped forward KL contributions. Our methods (\method{} and \methodplain{}) substantially change both patterns. \textbf{Left:} full $D_{\mathrm{KL}}(p_T\|p_S)$ over training. \textbf{Middle:} full $D_{\mathrm{KL}}(p_S\|p_T)$ over training. \textbf{Right:} $R_{10}$ (from step 20 to 100) for dominant mismatch directions. Error bars in all panels are sample standard deviations over 3 training seeds. }
    \label{fig:diagnostic}
\end{figure*}

\section{OPSD and its missing feedback path}

\subsection{Pointwise KL clipping in OPSD}
OPSD applies pointwise clipping to KL contributions~\citep{zhao2026opsd}. Let $\mathcal D=\{(x,r)\}$ be a dataset of problems $x$ paired with verified reasoning traces $r$, which are available only during training. We call $r$ the \textit{privileged context}. The student $p_S^\theta(\cdot\mid x)$ samples $\hat y$, while the privileged teacher $p_T(\cdot\mid x,r)$ scores each student prefix $h_n=(x,\hat y_{<n})$. We refer to $D_{\mathrm{KL}}(p_T\|p_S)$ as the \emph{forward KL}, where the teacher distribution is the target for student optimization, and to $D_{\mathrm{KL}}(p_S\|p_T)$ as the \emph{reverse KL}. At position $n$, the forward KL decomposes over vocabulary entries:
\begin{equation}
D_{\mathrm{KL}}(p_{T,n}\|p_{S,n})
=\sum_{v\in\mathcal V}c_{n,v}.
\label{eq:pointwise}
\end{equation}
\begin{equation}
c_{n,v}=p_{T,n}(v)\log\frac{p_{T,n}(v)}{p_{S,n}(v)}.
\end{equation}
Individual contributions can be negative. OPSD applies an upper clip to each vocabulary contribution before summation:
\begin{equation}
\widetilde D_{\mathrm{KL},\tau}(p_{T,n}\|p_{S,n})
=\sum_{v\in\mathcal V}\min(c_{n,v},\tau).
\label{eq:clip}
\end{equation}
When $c_{n,v}>\tau$, the capped term contributes no further gradient to the student.
Clipping remains active throughout training. In Qwen3-4B OPSD, about 56\% of the positive forward KL contribution mass is removed by the cap. The student therefore repeatedly suppresses a substantial part of the privileged signal, but the fixed teacher receives no information about which directions were suppressed.

\subsection{Persistent mismatch under OPSD}
\label{sec:fixed-diagnostic}
We examine whether the student diverges from the fixed privileged teacher during training and whether dominant mismatch directions recur on the same problem. We measure the distributional gap using both $D_{\mathrm{KL}}(p_T\|p_S)$ and $D_{\mathrm{KL}}(p_S\|p_T)$. To quantify recurrence, let $C_{i,t}^{(k)}$ denote the top-$k$ vocabulary entries ranked by the KL contributions at token position $i$ and checkpoint $t$. We report:
\begin{equation}
R_k(a,b)=\frac{1}{N}\sum_{i=1}^{N}
\frac{|C_{i,a}^{(k)}\cap C_{i,b}^{(k)}|}
{|C_{i,a}^{(k)}\cup C_{i,b}^{(k)}|},
\label{eq:recurrence-main}
\end{equation}
where $R_k=1$ means that the same dominant vocabulary directions persist at the two checkpoints and $R_k=0$ means that the selected sets do not overlap. Our primary statistic is $R_{10}$ between steps 20 and 100. Appendix~\ref{app:diagnostics} specifies the contribution cutoffs, protocol, coverage, and sensitivity to $k$.

Figure~\ref{fig:diagnostic} shows increasing mismatch and persistent clipped contributions under OPSD. Under OPSD, $D_{\mathrm{KL}}(p_T\|p_S)$ increases by roughly $11\times$ from the early to the final step, while $D_{\mathrm{KL}}(p_S\|p_T)$ grows by more than $2\times$. The dominant clipped forward KL directions are also highly persistent, with $R_{10}=0.693$ between steps 20 and 100. The mismatch therefore is not caused by random clipping. The teacher side output is repeatedly encountered on the same problems.

These observations motivate \method{}. We seek to reduce the avoidable distribution gap while preserving the advantage of privileged information. In our method, the student continues to learn on-policy from privileged supervision, and the teacher is updated using feedback from the student.

\section{\method{}: closing the loop while preserving privileged context}

\subsection{Asymmetric alternating update}

\method{} alternates between asymmetric student and teacher updates. We retain OPSD's clipping rule to limit the student's gradients from large privileged KL contributions~\citep{zhao2026opsd}. After the student update, the teacher learns from the updated student distributions on the same trajectory.

Let $\theta_k$ and $\phi_k$ denote the student and teacher parameters, and let $\sg[\cdot]$ denote the stop-gradient operator. The student first samples $\hat y_k\sim p_S^{\theta_k}(\cdot\mid x)$. On this trace, write $t_{n,v}=\sg[p_{T,n}^{\phi_k}(v)]$ and $s_{n,v}^{\theta}=p_{S,n}^{\theta}(v)$. The student retains OPSD's update, with the pointwise contribution and clipped token loss defined as
\begin{equation}
c_{n,v}^{\theta}
=t_{n,v}\log\frac{t_{n,v}}{s_{n,v}^{\theta}},
\end{equation}
\begin{equation}
\ell_{S,n}(\theta)=
\sum_{v\in\mathcal V}\min(c_{n,v}^{\theta},\tau).
\end{equation}
The trajectory-averaged student objective is
\begin{equation}
\mathcal L_S(\theta,\phi_k)
=\mathbb E\left[\frac{1}{|\hat y_k|}
\sum_n\ell_{S,n}(\theta)\right].
\label{eq:student}
\end{equation}

Optimizing Equation~\ref{eq:student} produces $\theta_{k+1}$. On the same trajectory, let $s_n^+=p_{S,n}^{\theta_{k+1}}$ be the updated student distribution and $s^+=(s_n^+)_n$ its sequence of token distributions. For any such sequence $u=(u_n)_n$, define the stopped-target teacher divergence and its trajectory average by

\begin{equation}
d_n(u_n,\phi)=D_{\mathrm{KL}}\!\left(\sg[u_n]\|p_{T,n}^{\phi}\right),
\end{equation}
\begin{equation}
\bar d(u,\phi)=\frac{1}{|\hat y_k|}\sum_n d_n(u_n,\phi).
\end{equation}

For ablation purposes, we provide a version of our method without the residual target, which is called \methodplain{}. The teacher loss of this method is described in Equation~\ref{eq:teacher-direct}:
\begin{equation}
\mathcal L_T^{\mathrm{direct}}(\phi\mid\theta_{k+1})
=\mathbb E[\bar d(s^+,\phi)].
\label{eq:teacher-direct}
\end{equation}

In our method, the teacher update follows the student update and reuses the same sampled trajectory, yielding the alternating order $\phi_k\rightarrow\theta_{k+1}\rightarrow\phi_{k+1}$. The updated teacher provides privileged supervision at the next optimization step.

\subsection{Preserving the effect of privileged context}

Closing the feedback loop creates a tension absent from OPSD: the more strongly the teacher is pulled toward an unprivileged student, the easier it is for the teacher to lose the privileged effect of $r$. We observe this effect in the privileged sensitivity diagnostic in Section~\ref{sec:privilege}. We therefore anchor the teacher target to the effect of privileged context on the frozen initial model $p^0$. 

For $h_n=(x,\hat y_{<n})$, we define the initial privileged log-probability residual
\begin{equation}
\Delta_n^0(v)=\log p^0(v\mid h_n,r)
-\log p^0(v\mid h_n).
\label{eq:residual}
\end{equation}
The residual measures how privileged context changes the frozen model's log probabilities. We add this residual, which is scaled by a fixed hyperparameter $\beta$, to define the teacher target:
\begin{equation}
a_{n,v}=\log s_n^+(v)+\beta\Delta_n^0(v),
\label{eq:target-score}
\end{equation}
\begin{equation}
q_n(v)=\operatorname{softmax}_{v\in\mathcal V}(a_{n,v}).
\label{eq:target}
\end{equation}

Writing $q=(q_n)_n$ for the sequence of residual targets, the \method{} teacher minimizes
\begin{equation}
\mathcal L_T(\phi\mid\theta_{k+1})
=\mathbb E[\bar d(q,\phi)].
\label{eq:teacher}
\end{equation}

The hyperparameter $\beta$ controls the residual strength relative to the updated student distribution. At $\beta=0$, Equation~\ref{eq:target-score} gives the updated student's log probabilities. The target then equals that student distribution, and Equation~\ref{eq:teacher} reduces to the loss in \methodplain{}. Increasing $\beta$ gives the frozen reference residual greater influence on the target, balancing adaptation to the current student against preservation of the initial effect of privileged context.

\subsection{Implementation and cost}
Algorithm~\ref{alg:dualopsd} summarizes one optimization step. We implement the student and teacher as independent LoRA adapters on the same frozen LLM backbone. Both objectives compute KL terms from full-vocabulary distributions. Only the student adapter is retained at inference time.

\begin{algorithm}[h]
\caption{DualOPSD (one optimization step)}
\label{alg:dualopsd}
\begin{algorithmic}[1]
\Require Batch $\mathcal B$, student $p_S^\theta$, teacher $p_T^\phi$, initial policy $p^0$, clip $\tau$, residual weight $\beta$
\For{$(x,r)\in\mathcal B$}
    \State Sample one response $\hat y\sim p_S^\theta(\cdot\mid x)$
    \State Score $\hat y$ with $p_S^\theta(\cdot\mid x)$ and $\sg[p_T^\phi(\cdot\mid x,r)]$
\EndFor
\State Update $\theta$ with the clipped student loss in Equation~\ref{eq:student}
\State \textcolor{blue}{Recompute $\sg[p_S^\theta]$ on the same trajectories}
\State \textcolor{blue}{Compute $\Delta^0$ and form the stopped target $q$ in Equation~\ref{eq:target}}
\State \textcolor{blue}{Update $\phi$ with the full teacher loss in Equation~\ref{eq:teacher}}
\State \Return updated student and teacher adapters
\end{algorithmic}
\end{algorithm}

The student update, residual construction, and teacher update reuse the same sampled response, so \method{} requires one rollout per problem, as in OPSD. The additional cost comes from extra full-vocabulary scoring and backpropagation, which increase the time for each step to approximately 1.8$\times$ that of OPSD.

\begin{table*}[t]

\centering
\small
\setlength{\tabcolsep}{4.5pt}
\begin{tabular}{llrrrr}
\toprule
Model & Method & AIME24 & AIME25 & HMMT25 & Mean \\
\midrule
\multirow{6}{*}{1.7B}
& Base & $11.94\pm1.94$ & $9.54\pm1.12$ & $5.74\pm0.42$ & $9.07\pm0.82$ \\
& SFT & $8.80\pm1.12$ & $7.87\pm0.64$ & $1.67\pm0.56$ & $6.11\pm0.65$ \\
& GRPO & $12.50\pm2.82$ & $\underline{9.72\pm0.96}$ & $4.81\pm0.32$ & $9.01\pm0.83$ \\
& OPSD$^{\ast}$ & $\underline{14.17\pm1.39}$ & $9.72\pm0.73$ & $\underline{6.48\pm1.25}$ & $\underline{10.12\pm0.62}$ \\
& \methodplain{} & $11.57\pm1.16$ & $8.52\pm1.05$ & $4.72\pm0.73$ & $8.27\pm0.93$ \\
& \method{} & $\mathbf{15.00\pm2.65}$ & $\mathbf{12.22\pm1.73}$ & $\mathbf{7.59\pm2.05}$ & $\mathbf{11.60\pm1.82}$ \\
\midrule
\multirow{6}{*}{4B}
& Base & $22.59\pm2.23$ & $20.09\pm2.31$ & $10.00\pm1.00$ & $17.56\pm1.84$ \\
& SFT & $19.81\pm1.70$ & $18.15\pm1.25$ & $9.72\pm1.47$ & $15.90\pm0.63$ \\
& GRPO & $22.31\pm0.85$ & $18.52\pm1.53$ & $11.30\pm1.12$ & $17.38\pm0.57$ \\
& OPSD$^{\ast}$ & $28.70\pm0.70$ & $24.81\pm1.12$ & $13.98\pm1.37$ & $22.50\pm0.82$ \\
& \methodplain{} & $\underline{39.07\pm6.51}$ & $\underline{30.28\pm1.21}$ & $\underline{20.00\pm3.61}$ & $\underline{29.78\pm3.75}$ \\
& \method{} & $\mathbf{42.78\pm1.21}$ & $\mathbf{37.41\pm4.18}$ & $\mathbf{24.26\pm2.05}$ & $\mathbf{34.81\pm1.80}$ \\
\midrule
\multirow{6}{*}{8B}
& Base & $26.39\pm1.27$ & $19.26\pm0.42$ & $11.02\pm0.70$ & $18.89\pm0.56$ \\
& SFT & $22.04\pm1.95$ & $16.67\pm1.00$ & $8.43\pm0.70$ & $15.71\pm0.62$ \\
& GRPO & $27.50\pm1.73$ & $19.35\pm1.12$ & $11.11\pm0.96$ & $19.32\pm0.35$ \\
& OPSD$^{\ast}$ & $37.22\pm2.90$ & $27.41\pm1.70$ & $15.37\pm1.12$ & $26.67\pm1.44$ \\
& \methodplain{} & $\underline{57.31\pm1.89}$ & $\underline{39.81\pm1.81}$ & $\mathbf{26.57\pm0.42}$ & $\underline{41.23\pm1.18}$ \\
& \method{} & $\mathbf{58.24\pm3.10}$ & $\mathbf{43.06\pm2.47}$ & $\underline{25.83\pm4.55}$ & $\mathbf{42.38\pm3.29}$ \\
\bottomrule
\end{tabular}
\caption{Avg@12 accuracy on three mathematics benchmarks (\%). Entries report means $\pm$ sample standard deviations across three seeds. The final column reports the mean accuracy across the three benchmarks. The 1.7B, 4B, and 8B residual variants use $\beta=0.8$, $\beta=0.7$, and $\beta=0.5$. $^\ast$ denotes that OPSD is reported at the more accurate of its 100-step and matched-compute checkpoints. The per-seed measurements underlying Table~\ref{tab:main} are reported in Appendix~\ref{app:seed}.}
\label{tab:main}
\end{table*}

\section{Experimental setup}
\label{sec:setup}

\paragraph{Data and models.}
We train on 29K mathematics problems and verified solutions from the OpenThoughts mathematics subset used by OPSD~\citep{guha2026openthoughts,zhao2026opsd}. We use Qwen3-1.7B, Qwen3-4B, and Qwen3-8B in non-thinking mode~\citep{yang2025qwen3}. We train the distillation methods for 100 steps with an effective batch size of 32 and student rollouts capped at 1,024 tokens. We use LoRA rank 64 with scale 128 and a learning rate of $5\times10^{-6}$ with 10 warmup steps. The pointwise student clipping threshold is $10^{-6}$ at 1.7B and 4B and $10^{-7}$ at 8B. Unless otherwise specified, results with uncertainty estimates report means and sample standard deviations across 3 training seeds.

\paragraph{Baselines.}
We compare \method{} with the untrained Base model, SFT on reference solutions, GRPO with verifiable rewards~\citep{shao2024deepseekmath}, OPSD~\citep{zhao2026opsd}, and \methodplain{}. \method{} uses $\beta$ selected on a held-out validation set, as detailed in Appendix~\ref{app:beta-selection}.
SFT is trained for 100 steps. OPSD is evaluated at both 100 steps and a scale-specific checkpoint selected to match the training compute of \method{}. GRPO is trained for 200 steps. All trained methods use the same data split and LoRA rank. Appendix~\ref{app:cost} defines the matching rule and reports the costs.

\paragraph{Evaluation.}
We evaluate all methods on three math benchmarks: AIME 2024, AIME 2025, and HMMT February 2025. We generate 12 responses per problem. The primary metric is avg@12, which is the mean accuracy. Decoding uses non-thinking mode, temperature 1.0, top-$p$ 1.0, and a 32,768-token total limit. The 1.7B, 4B, and 8B residual runs use the optimal $\beta$ selected on a held-out set, which contains 500 problems randomly selected from the original dataset.
Details are provided in Appendices~\ref{app:repro}, \ref{app:cost}, and~\ref{app:diagnostics}.

\section{Results and analysis}

\begin{table*}[t]
\centering
\small
\setlength{\tabcolsep}{2.5pt}
\begin{tabular}{llrrrrrr}
\toprule
Model & Method & PFLOPs & GPU h & $\Delta$ Acc. & $\Delta$ tokens & Trunc. (\%)$\downarrow$ & $E\uparrow$ \\
\midrule
\multirow{4}{*}{Overall}
& OPSD & $161.78$ & 1.00 & $+4.59$ & $+3.27\mathrm{K}$ & $6.45$ & 1.27 \\
& OPSD, matched compute & $436.52$ & 3.19 & $+2.89$ & $+4.66\mathrm{K}$ & $13.14$ & 0.59 \\
& \methodplain{} & $318.59$ & 1.40 & \underline{$+11.26$} & $+3.13\mathrm{K}$ & \textbf{2.42} & \underline{2.33} \\
& \method{} & $435.81$ & 1.78 & \textbf{+14.42} & $+5.51\mathrm{K}$ & \underline{$6.68$} & \textbf{2.57} \\
\bottomrule
\end{tabular}
\caption{Overall cost and generation behavior for distillation methods, averaged across the 1.7B, 4B, and 8B scales. $\Delta$ Acc. is the change in the three-benchmark avg@12 mean relative to the Base, in percentage points. $\Delta$ tokens is the change in mean observed completion tokens relative to Base. Length efficiency $E$ is defined in Equation~\ref{eq:length-efficiency}.}
\label{tab:tradeoff}
\end{table*}

\subsection{Main results}

\paragraph{The advantage of \method{} grows with model scale.}
Across training runs with 3 seeds and matched compute, \method{} consistently improves the mean accuracy over the three benchmarks at all model scales.
The gains were 1.48 percentage points at 1.7B, 12.31 at 4B, and 15.71 at 8B, indicating a larger advantage at larger scales. The ablation variant, \methodplain{}, showed a similar scaling trend. The residual target is more effective at 1.7B: \methodplain{} achieves 8.27 points, below the 10.12 points of OPSD. Incorporating the residual target allows \method{} to surpass OPSD.

\paragraph{SFT and GRPO yield limited gains under this training setup.}
SFT reduced mean accuracy relative to Base at all three model scales. GRPO remained close to Base but below OPSD and \method{}. The decline under SFT may reflect a mismatch between reference trajectories and the student's own output distribution, consistent with \citet{shenfeld2026sdft}. These comparisons support the benefit of privileged on-policy distillation.

\subsection{Compute and length tradeoff}

Response length can affect accuracy comparisons between reasoning methods~\citep{muennighoff2025s1}, and training compute differs across distillation methods. In this section we report training compute, measured GPU hours, observed completion length, and truncation rates. To summarize accuracy gains relative to changes in response length, we define response length efficiency
\begin{equation}
E(m)=\frac{\mathrm{Acc}(m)-\mathrm{Acc}(\mathrm{Base})}
{\bigl(\mathrm{Tok}(m)-\mathrm{Tok}(\mathrm{Base})\bigr)/1000},
\label{eq:length-efficiency}
\end{equation}
where accuracy is the avg@12 mean and $\mathrm{Tok}$ is the mean response length, including truncated responses. A larger $E$ indicates a greater accuracy improvement per additional 1K generated tokens relative to the Base model. For methods that generate longer responses than Base, a negative $E$ indicates an accuracy decline.

\paragraph{Our method improves overall response length efficiency.}
Both \method{} and \methodplain{} had higher overall length efficiency than OPSD (Table~\ref{tab:tradeoff}). At 4B, \methodplain{} and \method{} reach $3.7\times$ and $2.7\times$ the length efficiency of OPSD.
At 8B, both methods achieved approximately $2\times$ the efficiency of OPSD.
\methodplain{} had the lowest overall truncation rate at $2.42\%$, compared with $6.68\%$ for \method{} and $6.45\%$ for OPSD. The per-scale results are available in Table~\ref{tab:tradeoff-by-scale}.

\begin{figure}[h]
    \centering
    \includegraphics[width=0.8\columnwidth]{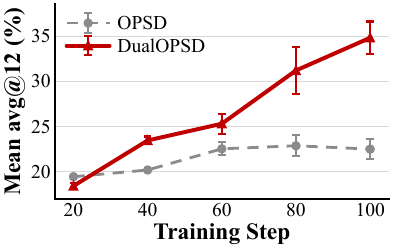}
    \caption{Qwen3-4B accuracy over training for OPSD and \method{}. Each curve shows the mean avg@12 across three benchmarks at 5 checkpoints from step 20 to 100.}
    \label{fig:acc-curves}
\end{figure}

\subsection{Effect of training steps}

\paragraph{\method{} continues to improve during training.}
To evaluate performance stability, we tested 5 training checkpoints from step 20 to step 100. As shown in Figure~\ref{fig:acc-curves}, on Qwen3-4B, the mean accuracy of \method{} increases from 18.43\% at step 20 to 34.81\% at step 100, while OPSD plateaus near 22.5\% after step 60. Additionally, we found that extending OPSD from 100 steps to the matched-compute checkpoints at approximately 250 steps increased its overall truncation rate by $2.03\times$ and also decreased the performance, which reveals the instability of OPSD.

\begin{figure*}[t]
    \centering
    \includegraphics[width=\textwidth]{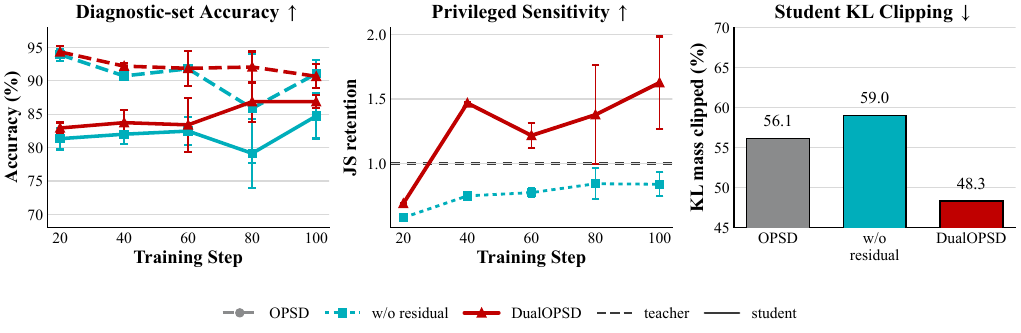}
    \caption{Teacher accuracy and privileged-reference sensitivity on Qwen3-4B. \textbf{Left:} teacher and student accuracy across five checkpoints on 500 diagnostic problems. \textbf{Middle:} normalized Jensen--Shannon (JS) retention. The horizontal line at one denotes the initial sensitivity level. \textbf{Right:} fraction of positive forward KL contribution mass removed by the student cap, averaged over the five checkpoints. OPSD appears only in the right panel.}
    \label{fig:ablation-dynamics}
\end{figure*}

\subsection{The effect of teacher adaptation}
\label{sec:privilege}

\paragraph{Teacher adaptation reduces mismatch while improving accuracy.}
On Qwen3-4B (Figure~\ref{fig:diagnostic}), $D_{\mathrm{KL}}(p_T\|p_S)$ decreases from $2.04$ under OPSD to $0.64$ with \methodplain{} and $0.76$ with \method{}. Reverse KL exhibits the same overall trend. The temporal structure of the mismatch also changes: the $R_{10}$ of clipped forward KL contributions drops from $0.69$ under OPSD to $0.36$ with \methodplain{} and $0.21$ with \method{}. Both of our methods substantially reduce bidirectional distribution mismatch relative to OPSD, and the residual target in \method{} further suppresses recurrence. This combination suggests that the residual design improves the temporal stability of teacher--student alignment.

\paragraph{The residual target preserves sensitivity to privileged context.}
In Figure~\ref{fig:ablation-dynamics} Left, we evaluated teacher and student responses on a separate diagnostic set of 500 problems, which has the same distribution as the training set. We found that \method{} had higher teacher and student accuracy than \methodplain{} on the diagnostic set. The teacher retained higher mean accuracy than the student at every checkpoint for both methods.

In Figure~\ref{fig:ablation-dynamics} Middle, we measured sensitivity to the privileged reference using the problem-level normalized Jensen--Shannon retention statistic, which is defined in Appendix~\ref{app:teacher-diagnostics}. We found that \method{} has nearly twice the sensitivity of \methodplain{} (1.626 vs.\ 0.838). Thus the residual target helps to keep sensitivity during teacher updates.

In Figure~\ref{fig:ablation-dynamics} Right, the student clipped mass in our method also decreases. The clipped mass drops from 56.1\% in OPSD to 48.3\% in \method{}. We also found that \methodplain{} has higher clipped mass than \method{}, which indicates that the residual target also helps to reduce misalignment between student and teacher.

\subsection{Ablation study on teacher KL clipping}

\begin{table}[h]

\centering
\small
\setlength{\tabcolsep}{1.6pt}
\begin{tabular}{lccc}
\toprule
Metric & High Clip & Low Clip & No Clip \\
\midrule
Removed KL & Upper & Lower & None \\
\midrule
\multicolumn{4}{l}{\textit{Qwen3-1.7B}} \\
Acc. (\%) & $7.59\pm0.48$ & $9.88\pm0.81$ & $\mathbf{11.60\pm1.82}$ \\
Len. (K) & $2.13\pm0.16$ & $4.40\pm0.67$ & $5.22\pm0.96$ \\
Trunc. (\%) & $1.23\pm0.33$ & $6.42\pm0.63$ & $8.77\pm3.23$ \\
\midrule
\multicolumn{4}{l}{\textit{Qwen3-4B}} \\
Acc. (\%) & $22.16\pm1.72$ & $28.46\pm4.88$ & $\mathbf{34.81\pm1.80}$ \\
Len. (K) & $3.16\pm0.67$ & $5.83\pm1.64$ & $6.81\pm1.16$ \\
Trunc. (\%) & $5.96\pm1.68$ & $8.09\pm4.15$ & $5.68\pm2.52$ \\
\midrule
\multicolumn{4}{l}{\textit{Qwen3-8B}} \\
Acc. (\%) & $23.67\pm2.94$ & $\mathbf{42.59\pm2.94}$ & $42.38\pm3.29$ \\
Len. (K) & $2.38\pm0.31$ & $8.12\pm0.40$ & $7.39\pm1.05$ \\
Trunc. (\%) & $4.10\pm0.84$ & $4.51\pm2.59$ & $5.59\pm2.53$ \\
\bottomrule
\end{tabular}
\caption{Teacher-side clipping ablation on the \method{} teacher objective at each Qwen3 scale. Acc. is the mean avg@12 across the three benchmarks, Len. is the mean length of non-truncated responses in K tokens, and Trunc. is the truncation rate. Entries report means $\pm$ sample standard deviations across three training seeds.}
\label{tab:core-ablation}
\end{table}

We investigate the effect of clipping on the teacher side. Table~\ref{tab:core-ablation} compares three clipping rules applied to the \method{} teacher objective $D_{\mathrm{KL}}(\sg[q]\|p_T)$. No clipping gave the highest accuracy at 1.7B and 4B, and was within $0.21$ percentage points of low clipping at 8B. Averaged over the three scales, no clipping yielded $29.60\%$ mean accuracy, compared with $26.98\%$ for low clipping and $17.81\%$ for high clipping. Because no clipping performed best on average, we use it for the teacher update in \method{}.
Appendix~\ref{app:teacher-objectives} gives the clipping rules and results on each benchmark.

\subsection{Teacher--student style alignment}

OPSD clips large KL contributions in part because the privileged teacher writes in a style that differs from the student's. We therefore test whether teacher adaptation reduces this difference directly. For each problem, the teacher and student generate responses with the same seed, and we compare nine formatting features, such as newlines, Markdown headings, bullets, bold spans, and display mathematics. The distance is the mean absolute standardized difference (Equation~\ref{eq:style-distance}).

\begin{table}[h]

\centering
\small
\setlength{\tabcolsep}{3pt}
\begin{tabular}{lccc}
\toprule
Model & OPSD & \method{} & Reduction (\%) \\
\midrule
1.7B & $0.823\pm0.022$ & $\mathbf{0.467\pm0.049}$ & 43.2 \\
4B   & $0.810\pm0.006$ & $\mathbf{0.266\pm0.069}$ & 67.2 \\
8B   & $0.830\pm0.007$ & $\mathbf{0.390\pm0.039}$ & 53.0 \\
\bottomrule
\end{tabular}
\caption{Teacher--student formatting distance at step 100 on 500 diagnostic problems. Lower values indicate closer agreement.}
\label{tab:style-alignment-main}
\end{table}

\paragraph{\method{} reduces the formatting distance.} Table~\ref{tab:style-alignment-main} shows that our method reduces the distance relative to OPSD at all three scales, by 43.2\%, 67.2\%, and 53.0\% at 1.7B, 4B, and 8B. The reduction holds in each experiment. These results support the view that the adapted teacher produces responses closer to student style, but the fixed OPSD teacher keeps its own formatting.
Appendix~\ref{app:style-alignment} gives the protocol and qualitative examples.

\section{Related Work}

\paragraph{Off-policy and on-policy distillation.}
Knowledge distillation refers to learning from teacher predictions and is used to transfer model behavior~\citep{hinton2015distilling}.
Approaches to sequence distillation can be organized according to the source of training trajectories.
Off-policy methods rely on predefined teacher completions or rationales to supervise the student, whereas on-policy methods use student generated prefixes~\citep{kim2016sequence,hsieh2023stepbystep,agarwal2024onpolicy}.
For on-policy distillation, MiniLLM uses reverse-KL matching to transfer teacher behavior, whereas GKD uses alternative divergence objectives on student generated sequences~\citep{gu2023minillm,agarwal2024onpolicy}.
Within this area, our work focuses on student generated trajectories and examines how their supervision changes when the teacher adapts.

\paragraph{Self-distillation.}
Approaches to self-distillation can be organized according to how their targets are constructed.
One line of work uses previous predictions or policy samples ranked by reward to construct supervision~\citep{furlanello2018bornagain,sessa2025bond}.
OPSD achieves privileged self-supervision by conditioning a copy of the model on verified solutions at student-generated prefixes~\citep{zhao2026opsd}.
Our question concerns teacher adaptation within privileged self-distillation, specifically how to retain the reference signal as the student changes.

\paragraph{Dynamic teachers.}
Research on dynamic teachers investigates how to adapt supervision during student learning.
Some studies use parameter averaging or mutual prediction matching to update teachers~\citep{tarvainen2017meanteacher,zhang2018deepmutual}.
Our work is closely related to PAST and VISTA, which also adapt privileged teachers using student information~\citep{feng2026past,ding2026vista}.
The methods differ in adaptation signals: PAST uses complete trajectories and outcome-conditioned objectives, while our method uses updated student distributions with a frozen privileged residual.
Although both VISTA and \method{} have student feedback, they condition teacher updates on different information. VISTA uses verified positions with high disagreement as feedback, while our method uses updated student distributions with a frozen privileged residual.

\section{Conclusion}

\method{} closes OPSD's feedback loop through alternating student and teacher updates. Across Qwen3-1.7B, 4B, and 8B, it improves mean avg@12 over three benchmarks by 1.48, 12.31, and 15.71 percentage points relative to OPSD.
The ablations show that teacher adaptation reduces distributional mismatch and its recurrence. We also found that the residual target preserves privileged sensitivity.
Together, these results highlight the effectiveness of adapting the privileged teacher alongside the student in on-policy self-distillation. Future research can focus on teacher clipping strategies or other methods to further enhance teacher-student alignment.

\section*{Limitations}

Our evaluation covers Qwen3 models in non-thinking mode on mathematical reasoning benchmarks. We did not test other domains or privileged context in different formats. 
Although we found that our method has better performance than OPSD under an equal total time/compute budget, our method uses more time per training step, which is approximately $1.8\times$ longer than OPSD.

\section*{Ethical Considerations}
\label{sec:ethics}

This work studies self-distillation for mathematical reasoning using public datasets and pretrained Qwen3 models. It involves no human participants or annotators and does not collect new personal data. We use the datasets, models, and software for non-commercial research. 

\method{} may inherit errors, biases, and harmful behaviors from the pretrained backbone and reference solutions. Teacher adaptation and self-distillation may also reinforce undesirable behaviors, even when mathematical accuracy improves. Our evaluation focuses on mathematical reasoning and does not assess harmful output generation, fairness, or safety alignment. The reported accuracy gains and teacher--student alignment diagnostics therefore do not establish safety or reliability in other applications. Models adapted with this method should undergo safety evaluation before deployment.

Training and inference also consume computational resources. \method{} introduces additional scoring and teacher updates, and improved accuracy does not by itself imply a lower environmental impact. We report training compute and GPU hours in Appendix~\ref{app:cost} to make these costs transparent.
\bibliography{references}

@article{hinton2015distilling,
  title={Distilling the Knowledge in a Neural Network},
  author={Hinton, Geoffrey and Vinyals, Oriol and Dean, Jeff},
  journal={arXiv preprint arXiv:1503.02531},
  year={2015}
}

@inproceedings{kim2016sequence,
  title={Sequence-Level Knowledge Distillation},
  author={Kim, Yoon and Rush, Alexander M.},
  booktitle={Proceedings of EMNLP},
  pages={1317--1327},
  year={2016},
  doi={10.18653/v1/D16-1139},
  url={https://aclanthology.org/D16-1139/}
}

@inproceedings{gu2023minillm,
  title={{MiniLLM}: Knowledge Distillation of Large Language Models},
  author={Gu, Yuxian and Dong, Li and Wei, Furu and Huang, Minlie},
  booktitle={International Conference on Learning Representations},
  year={2024}
}

@inproceedings{agarwal2024onpolicy,
  title={On-Policy Distillation of Language Models: Learning from Self-Generated Mistakes},
  author={Agarwal, Rishabh and Vieillard, Nino and Zhou, Yongchao and Stanczyk, Piotr and Ramos, Sabela and Geist, Matthieu and Bachem, Olivier},
  booktitle={International Conference on Learning Representations},
  year={2024}
}

@article{shao2024deepseekmath,
  title={{DeepSeekMath}: Pushing the Limits of Mathematical Reasoning in Open Language Models},
  author={Shao, Zhihong and Wang, Peiyi and Zhu, Qihao and Xu, Runxin and Song, Junxiao and Bi, Xiao and Zhang, Haowei and Zhang, Mingchuan and Li, Y. K. and Wu, Y. and Guo, Daya},
  journal={arXiv preprint arXiv:2402.03300v3},
  year={2024},
  url={https://arxiv.org/abs/2402.03300v3}
}

@article{schulman2017ppo,
  title={Proximal Policy Optimization Algorithms},
  author={Schulman, John and Wolski, Filip and Dhariwal, Prafulla and Radford, Alec and Klimov, Oleg},
  journal={arXiv preprint arXiv:1707.06347},
  year={2017}
}

@article{guo2025deepseekr1,
  title={{DeepSeek-R1} Incentivizes Reasoning in {LLMs} through Reinforcement Learning},
  author={Guo, Daya and Yang, Dejian and Zhang, Haowei and Song, Junxiao and Wang, Peiyi and Zhu, Qihao and Xu, Runxin and Zhang, Ruoyu and Ma, Shirong and Bi, Xiao and others},
  journal={Nature},
  volume={645},
  number={8081},
  pages={633--638},
  year={2025},
  doi={10.1038/s41586-025-09422-z}
}

@inproceedings{zhao2026opsd,
  title={Self-Distilled Reasoner: On-Policy Self-Distillation for Large Language Models},
  author={Zhao, Siyan and Xie, Zhihui and Liu, Mengchen and Huang, Jing and Pang, Guan and Chen, Feiyu and Grover, Aditya},
  booktitle={Proceedings of the 43rd International Conference on Machine Learning},
  series={Proceedings of Machine Learning Research},
  volume={306},
  pages={162433--162448},
  publisher={PMLR},
  year={2026}
}

@article{yang2025qwen3,
  title={{Qwen3} Technical Report},
  author={Yang, An and Li, Anfeng and Yang, Baosong and Zhang, Beichen and Hui, Binyuan and Zheng, Bo and Yu, Bowen and Gao, Chang and Huang, Chengen and Lv, Chenxu and others},
  journal={arXiv preprint arXiv:2505.09388},
  year={2025}
}

@inproceedings{guha2026openthoughts,
  title={{OpenThoughts}: Data Recipes for Reasoning Models},
  author={Guha, Etash and Marten, Ryan and Keh, Sedrick and Raoof, Negin and Smyrnis, Georgios and Bansal, Hritik and Nezhurina, Marianna and Mercat, Jean and Vu, Trung and Sprague, Zayne and others},
  booktitle={International Conference on Learning Representations},
  year={2026}
}

@inproceedings{zhang2018deepmutual,
  title={Deep Mutual Learning},
  author={Zhang, Ying and Xiang, Tao and Hospedales, Timothy M. and Lu, Huchuan},
  booktitle={Proceedings of the IEEE Conference on Computer Vision and Pattern Recognition},
  pages={4320--4328},
  year={2018},
  doi={10.1109/CVPR.2018.00454}
}

@inproceedings{hsieh2023stepbystep,
  title={Distilling Step-by-Step! Outperforming Larger Language Models with Less Training Data and Smaller Model Sizes},
  author={Hsieh, Cheng-Yu and Li, Chun-Liang and Yeh, Chih-Kuan and Nakhost, Hootan and Fujii, Yasuhisa and Ratner, Alex and Krishna, Ranjay and Lee, Chen-Yu and Pfister, Tomas},
  booktitle={Findings of the Association for Computational Linguistics: ACL 2023},
  pages={8003--8017},
  year={2023},
  doi={10.18653/v1/2023.findings-acl.507},
  url={https://aclanthology.org/2023.findings-acl.507/}
}

@inproceedings{ko2024distillm,
  title={{DistiLLM}: Towards Streamlined Distillation for Large Language Models},
  author={Ko, Jongwoo and Kim, Sungnyun and Chen, Tianyi and Yun, Se-Young},
  booktitle={International Conference on Machine Learning},
  year={2024},
  volume={235},
  pages={24872--24895},
  url={https://proceedings.mlr.press/v235/ko24c.html}
}

@inproceedings{furlanello2018bornagain,
  title={Born Again Neural Networks},
  author={Furlanello, Tommaso and Lipton, Zachary C. and Tschannen, Michael and Itti, Laurent and Anandkumar, Anima},
  booktitle={International Conference on Machine Learning},
  pages={1607--1616},
  year={2018},
  url={https://proceedings.mlr.press/v80/furlanello18a.html}
}

@inproceedings{anil2018codistillation,
  title={Large Scale Distributed Neural Network Training Through Online Distillation},
  author={Anil, Rohan and Pereyra, Gabriel and Passos, Alexandre and Ormandi, Robert and Dahl, George E. and Hinton, Geoffrey E.},
  booktitle={International Conference on Learning Representations},
  year={2018}
}

@inproceedings{tarvainen2017meanteacher,
  title={Mean Teachers Are Better Role Models: Weight-Averaged Consistency Targets Improve Semi-Supervised Deep Learning Results},
  author={Tarvainen, Antti and Valpola, Harri},
  booktitle={Advances in Neural Information Processing Systems},
  pages={1195--1204},
  year={2017}
}

@inproceedings{rafailov2023dpo,
  title={Direct Preference Optimization: Your Language Model is Secretly a Reward Model},
  author={Rafailov, Rafael and Sharma, Archit and Mitchell, Eric and Manning, Christopher D. and Ermon, Stefano and Finn, Chelsea},
  booktitle={Advances in Neural Information Processing Systems},
  year={2023},
  doi={10.52202/075280-2338},
  url={https://proceedings.neurips.cc/paper_files/paper/2023/hash/a85b405ed65c6477a4fe8302b5e06ce7-Abstract-Conference.html}
}

@inproceedings{jang2026veto,
  title={Stable On-Policy Distillation through Adaptive Target Reformulation},
  author={Jang, Ijun and Yeom, Jewon and Yeo, Juan and Lim, Hyunggyu and Kim, Taesup},
  booktitle={Findings of the Association for Computational Linguistics: ACL 2026},
  pages={42217--42227},
  year={2026},
  doi={10.18653/v1/2026.findings-acl.2094},
  url={https://aclanthology.org/2026.findings-acl.2094/}
}

@article{ko2026reopold,
  title={Scaling Reasoning Efficiently via Relaxed On-Policy Distillation},
  author={Ko, Jongwoo and Abdali, Sara and Kim, Young Jin and Chen, Tianyi and Cameron, Pashmina},
  journal={arXiv preprint arXiv:2603.11137},
  year={2026}
}

@article{liu2026pwopsd,
  title={When Are Teacher Tokens Reliable? Position-Weighted On-Policy Self-Distillation for Reasoning},
  author={Liu, Xiaogeng and Wang, Xinyan and Ma, Yingzi and Zhang, Yechao and Xiao, Chaowei},
  journal={arXiv preprint arXiv:2605.21606},
  year={2026}
}

@article{wang2026teachability,
  title={Not All Disagreement Is Learnable: Token Teachability in On-Policy Distillation},
  author={Wang, Yuanyi and Lu, Su and Gu, Yanggan and Wang, Pengkai and Yang, Yifan and Yan, Zhaoyi and Xie, Congkai and Wu, Jianmin and Yang, Hongxia},
  journal={arXiv preprint arXiv:2605.26844},
  year={2026}
}

@article{hou2026uniopd,
  title={{Uni-OPD}: Unifying On-Policy Distillation with a Dual-Perspective Recipe},
  author={Hou, Wenjin and Peng, Shangpin and Wang, Weinong and Ruan, Zheng and Zhang, Yue and Zhou, Zhenglin and Gao, Mingqi and Chen, Yifei and Wang, Kaiqi and Yang, Hongming and Zhang, Chengquan and Tian, Zhuotao and Hu, Han and Yang, Yi and Wu, Fei and Fan, Hehe},
  journal={arXiv preprint arXiv:2605.03677},
  year={2026}
}

@article{han2026atesd,
  title={Adaptive Teacher Exposure for Self-Distillation in LLM Reasoning},
  author={Han, Zihao and Zhang, Tiangang and Wang, Huaibin and Sun, Yilun},
  journal={arXiv preprint arXiv:2605.11458},
  year={2026}
}

@article{yu2026dopd,
  title={{DOPD}: Dual On-policy Distillation},
  author={Yu, Xinlei and Li, Gen and Si, Qingyi and Zhang, Guibin and Xu, Yuqi and Wang, Congcong and Dong, Shuai and Tuo, Kaiwen and Zeng, Xiangyu and Feng, Kaituo and Wang, Qunzhong and Shi, Yang and Hu, Xiaobin and Yue, Xiangyu and Wang, Jiaqi and Yan, Shuicheng},
  journal={arXiv preprint arXiv:2606.30626},
  year={2026}
}

@article{feng2026past,
  title={{PAST}: Privileged Adaptation from Complete Student Trajectories for On-Policy Self-Distillation},
  author={Feng, Yangyang and Feng, Zhuoyan and Chen, Junlan},
  journal={arXiv preprint arXiv:2608.08726},
  year={2026}
}

@inproceedings{sessa2025bond,
  title={{BOND}: Aligning {LLM}s with Best-of-{N} Distillation},
  author={Sessa, Pier Giuseppe and Dadashi, Robert and Hussenot-Desenonges, L{\'e}onard and Ferret, Johan and Vieillard, Nino and Rame, Alexandre and Shahriari, Bobak and Perrin, Sarah and Friesen, Abram and Cideron, Geoffrey and Girgin, Sertan and Stanczyk, Piotr and Michi, Andrea and Sinopalnikov, Danila and Ramos Garea, Sabela and H{\'e}liou, Am{\'e}lie and Severyn, Aliaksei and Hoffman, Matthew and Momchev, Nikola and Bachem, Olivier},
  booktitle={International Conference on Learning Representations},
  year={2025}
}

@inproceedings{sprague2026skillfactory,
  title={{SkillFactory}: Self-Distillation for Learning Cognitive Behaviors},
  author={Sprague, Zayne and Lu, Jack and Wadhwa, Manya and Keh, Sedrick and Ren, Mengye and Durrett, Greg},
  booktitle={International Conference on Learning Representations},
  year={2026}
}

@inproceedings{yang2025supercorrect,
  title={{SuperCorrect}: Advancing Small {LLM} Reasoning with Thought Template Distillation and Self-Correction},
  author={Yang, Ling and Yu, Zhaochen and Zhang, Tianjun and Xu, Minkai and Gonzalez, Joseph E. and Cui, Bin and Yan, Shuicheng},
  booktitle={International Conference on Learning Representations},
  year={2025}
}

@inproceedings{wen2026scaffolding,
  title={Reasoning Scaffolding: Distilling the Flow of Thought from {LLM}s},
  author={Wen, Xiangyu and Huang, Junhua and Li, Zeju and Li, Min and Zhong, Jianyuan and Xu, Zhijian and Yuan, Mingxuan and Huang, Yongxiang and Xu, Qiang},
  booktitle={International Conference on Learning Representations},
  year={2026}
}

@article{nie2026states,
  title={Post-Training is About States, Not Tokens: A State Distribution View of {SFT}, {RL}, and On-Policy Distillation},
  author={Nie, Dong},
  journal={arXiv preprint arXiv:2605.22731},
  year={2026},
  url={https://arxiv.org/abs/2605.22731}
}

@article{li2026rethinkingopd,
  title={Rethinking On-Policy Distillation of Large Language Models: Phenomenology, Mechanism, and Recipe},
  author={Li, Yaxuan and Zuo, Yuxin and He, Bingxiang and Zhang, Jinqian and Xiao, Chaojun and Qian, Cheng and Yu, Tianyu and Gao, Huan-ang and Yang, Wenkai and Liu, Zhiyuan and Ding, Ning},
  journal={arXiv preprint arXiv:2604.13016},
  year={2026},
  url={https://arxiv.org/abs/2604.13016}
}

@inproceedings{shenfeld2026sdft,
  title={Self-Distillation Enables Continual Learning},
  author={Shenfeld, Idan and Damani, Mehul and H{\"u}botter, Jonas and Agrawal, Pulkit},
  booktitle={Proceedings of the 43rd International Conference on Machine Learning},
  series={Proceedings of Machine Learning Research},
  volume={306},
  pages={111542--111558},
  publisher={PMLR},
  year={2026}
}

@inproceedings{muennighoff2025s1,
  title={{s1}: Simple Test-Time Scaling},
  author={Muennighoff, Niklas and Yang, Zitong and Shi, Weijia and Li, Xiang Lisa and Fei-Fei, Li and Hajishirzi, Hannaneh and Zettlemoyer, Luke and Liang, Percy and Cand{\`e}s, Emmanuel and Hashimoto, Tatsunori},
  booktitle={Proceedings of the 2025 Conference on Empirical Methods in Natural Language Processing},
  pages={20275--20321},
  publisher={Association for Computational Linguistics},
  year={2025},
  doi={10.18653/v1/2025.emnlp-main.1025},
  url={https://aclanthology.org/2025.emnlp-main.1025/}
}

@article{ye2026opcd,
  title={On-Policy Context Distillation for Language Models},
  author={Ye, Tianzhu and Dong, Li and Wu, Xun and Huang, Shaohan and Wei, Furu},
  journal={arXiv preprint arXiv:2602.12275},
  year={2026},
  url={https://arxiv.org/abs/2602.12275}
}

@article{fu2026revisitingopd,
  title={Revisiting On-Policy Distillation: Empirical Failure Modes and Simple Fixes},
  author={Fu, Yuqian and Huang, Haohuan and Jiang, Kaiwen and Liu, Jiacai and Jiang, Zhuo and Zhu, Yuanheng and Zhao, Dongbin},
  journal={arXiv preprint arXiv:2603.25562},
  year={2026},
  url={https://arxiv.org/abs/2603.25562}
}

@article{xu2026tip,
  title={{TIP}: Token Importance in On-Policy Distillation},
  author={Xu, Yuanda and Sang, Hejian and Zhou, Zhengze and He, Ran and Wang, Zhipeng and Geramifard, Alborz},
  journal={arXiv preprint arXiv:2604.14084},
  year={2026},
  url={https://arxiv.org/abs/2604.14084}
}

@article{jia2026aopd,
  title={Asymmetric On-Policy Distillation: Bridging Exploitation and Imitation at the Token Level},
  author={Jia, Nan and Yang, Haojin and Ma, Xing and Lian, Jiesong and Zhang, Shuailiang and Zhang, Weipeng and Zeng, Ke and Cai, Xunliang and Sun, Zequn},
  journal={arXiv preprint arXiv:2605.06387},
  year={2026},
  url={https://arxiv.org/abs/2605.06387}
}

@article{akhondzadeh2026rewardgated,
  title={Reward-Gated On-Policy Distillation},
  author={Akhondzadeh, Mohammad Sadegh and Lingam, Vijay and Tejaswi, Atula and Ekbote, Chanakya and Sanghavi, Sujay and Bojchevski, Aleksandar},
  journal={arXiv preprint arXiv:2607.04037},
  year={2026},
  url={https://arxiv.org/abs/2607.04037}
}

@article{xu2026betaopsd,
  title={{$\beta$-OPSD}: Deriving with Policy Optimization, Training with Self-Distillation},
  author={Xu, Jiawei and Liu, Minghui and Zhang, Juzheng and Goldstein, Tom and Huang, Furong},
  journal={arXiv preprint arXiv:2607.28582},
  year={2026},
  url={https://arxiv.org/abs/2607.28582}
}

@article{yu2026tide,
  title={Mismatch Matters: On-Policy Distillation Beyond Token Agreement},
  author={Yu, Zichao and Yu, Chengzhi and Xu, Shengze and Han, Yujin and Jiang, Bingqing and Wang, Xu and Zou, Difan},
  journal={arXiv preprint arXiv:2608.09836},
  year={2026},
  url={https://arxiv.org/abs/2608.09836}
}

@inproceedings{penaloza2026pidistill,
  title={Privileged Information Distillation for Language Models},
  author={Penaloza, Emiliano and Vattikonda, Dheeraj and Gontier, Nicolas and Lacoste, Alexandre and Charlin, Laurent and Caccia, Massimo},
  booktitle={Proceedings of the 43rd International Conference on Machine Learning},
  series={Proceedings of Machine Learning Research},
  volume={306},
  pages={97829--97868},
  publisher={PMLR},
  year={2026}
}

@article{ding2026vista,
  title={{VISTA}: Verifier-Informed Student-to-Teacher Adaptation for On-Policy Self-Distillation},
  author={Ding, Zewen and Wu, Zezhong and Tao, Zhou and Wang, Shida and Hou, Shizhuo and Hua, YongXiang and Cao, Haoyu and Xu, Linli},
  journal={arXiv preprint arXiv:2608.28306},
  year={2026},
  url={https://arxiv.org/abs/2608.28306}
}

@article{feng2026directopd,
  title={Weak-to-Strong Generalization via Direct On-Policy Distillation},
  author={Feng, Shiyuan and Gao, Huan-ang and Chi, Haohan and Wu, Hanlin and Zhang, Zhilong and Jiang, Zheng and He, Bingxiang and Ma, Wei-Ying and Zhang, Ya-Qin and Zhou, Hao},
  journal={arXiv preprint arXiv:2607.05394},
  year={2026},
  url={https://arxiv.org/abs/2607.05394}
}

@article{zhang2026aropd,
  title={Beyond Absolute Imitation: Anchored Residual Guidance for Privileged On-Policy Distillation},
  author={Zhang, Wenhao},
  journal={arXiv preprint arXiv:2606.10385},
  year={2026},
  url={https://arxiv.org/abs/2606.10385}
}

@article{nguyen2026avsd,
  title={{AVSD}: Adaptive-View Self-Distillation by Balancing Consensus and Teacher-Specific Privileged Signals},
  author={Nguyen, Duy and Xiao, Hanqi and Prasad, Archiki and Khan, Zaid and Das, Anirban and Zhang, Austin and Sahu, Sambit and Lee, Hyunji and Stengel-Eskin, Elias and Bansal, Mohit},
  journal={arXiv preprint arXiv:2605.20643},
  year={2026},
  url={https://arxiv.org/abs/2605.20643}
}

@article{li2026fireopd,
  title={Filter, Then Reweight: Rethinking Optimization Granularity in On-Policy Distillation},
  author={Li, Yuying and Zheng, Leqi and Yu, Yongzi and Zhou, Wenrui and Zhong, Xuchang and Hu, Xing and Jin, Jing and Yuan, Hangjie and Feng, Tao},
  journal={arXiv preprint arXiv:2606.02684},
  year={2026},
  url={https://arxiv.org/abs/2606.02684}
}

@article{zheng2026scope,
  title={{SCOPE}: Signal-Calibrated On-Policy Distillation Enhancement with Dual-Path Adaptive Weighting},
  author={Zheng, Binbin and Ma, Xing and Liang, Yiheng and Ruan, Jingqing and Fu, Xiaoliang and Lin, Kepeng and Zhu, Benchang and Zeng, Ke and Cai, Xunliang},
  journal={arXiv preprint arXiv:2604.10688},
  year={2026},
  url={https://arxiv.org/abs/2604.10688}
}

@article{yu2026mopd,
  title={Multi-Rollout On-Policy Distillation via Peer Successes and Failures},
  author={Yu, Weichen and Li, Xiaomin and Zhao, Yizhou and Liu, Xiaoze and Zhang, Ruowang and Wang, Haixin and Luo, Yinyi and Wu, Chen Henry and Mittal, Gaurav and Fredrikson, Matt and Hu, Yu},
  journal={arXiv preprint arXiv:2605.12652},
  year={2026},
  url={https://arxiv.org/abs/2605.12652}
}

@article{gkountouras2026canon,
  title={Consensus as Privileged Context for Label-Free Self-Distillation},
  author={Gkountouras, John and Juki{\'c}, Josip and Titov, Ivan},
  journal={arXiv preprint arXiv:2607.13643},
  year={2026},
  url={https://arxiv.org/abs/2607.13643}
}

\appendix

\section{Additional related work}
\label{app:related-work}

\paragraph{Distillation methods.}
Distillation methods differ in the source and form of teacher supervision. Some learn directly from teacher output distributions or generated responses.
\citet{kim2016sequence} train students on teacher generated completions. Born-Again Networks use a previous generation of the same model as the teacher and minimize cross-entropy against its soft predictions~\citep{furlanello2018bornagain}.  
Other methods combine teacher supervision with reinforcement learning.
SuperCorrect~\citep{yang2025supercorrect} first uses supervised finetuning on teacher generated thought templates, then uses Direct Preference Optimization (DPO)~\citep{rafailov2023dpo} to prefer corrected traces over the student's own corrections.
SkillFactory first finetunes the student on self generated traces restructured to exhibit cognitive skills such as verification and retry, then applies GRPO with outcome rewards~\citep{sprague2026skillfactory}.

Distillation can also use auxiliary training objectives.
Reasoning Scaffolding combines teacher distillation with an auxiliary objective that predicts reasoning signals for subsequent steps~\citep{wen2026scaffolding}. Similarly, \citet{hsieh2023stepbystep} train the student to predict answer labels while also minimizing cross-entropy on teacher generated explanation tokens.

\paragraph{On-policy distillation.}
On-policy distillation uses teacher feedback on student generated trajectories. For example, \citet{ye2026opcd} minimize reverse KL between the student and teacher on student generated trajectories.
Some methods modify the divergence objective used to compare teacher and student. \citet{ko2024distillm} minimize skew KL while reusing student samples for distillation. \citet{agarwal2024onpolicy} study different divergence objectives on student generated trajectories. \citet{fu2026revisitingopd} train students to match the teacher over a truncated top-$K$ support at student generated prefixes.

Other methods modify the target distribution or token-level loss to reduce harmful gradients or mitigate imbalanced predictions.
\citet{jang2026veto} train students against an intermediate distribution designed to suppress harmful gradients. $\beta$-OPSD trains students to match a geometric mixture of a reference policy and a privileged teacher~\citep{xu2026betaopsd}. TIDE uses a bounded Hellinger loss for overpredicted tokens and teacher top-$K$ guidance for underpredicted tokens~\citep{yu2026tide}.

Trajectory and token selection provide another way to focus distillation on useful supervision.
TIP applies distillation only at positions selected according to student entropy and teacher disagreement~\citep{xu2026tip}. \citet{liu2026pwopsd} weight the clipped forward KL loss according to the reliability of teacher guidance at each position. \citet{wang2026teachability} apply distillation where the student is likely to benefit from the teacher's guidance. FiRe-OPD filters poor trajectories and assigns larger weights to informative tokens in the remaining distillation loss~\citep{li2026fireopd}. SCOPE applies teacher-guided KL to incorrect rollouts and maximum-likelihood training to correct rollouts~\citep{zheng2026scope}.

\begin{table*}[!t]

\centering
\small
\begin{tabular}{llccc}
\toprule
Signal & Setting & OPSD & \methodplain{} & \method{} \\
\midrule
\multirow{5}{*}{Clipped forward KL}
 & \(k=1\)  & \textbf{0.720\ensuremath{\pm}0.005} & \underline{0.445\ensuremath{\pm}0.008} & 0.314\ensuremath{\pm}0.031 \\
 & \(k=5\)  & \textbf{0.698\ensuremath{\pm}0.005} & \underline{0.373\ensuremath{\pm}0.008} & 0.235\ensuremath{\pm}0.027 \\
 & \(k=10\) & \textbf{0.693\ensuremath{\pm}0.005} & \underline{0.355\ensuremath{\pm}0.010} & 0.208\ensuremath{\pm}0.028 \\
 & \(k=20\) & \textbf{0.682\ensuremath{\pm}0.005} & \underline{0.338\ensuremath{\pm}0.012} & 0.182\ensuremath{\pm}0.028 \\
 & Random-pair & \underline{0.0101\ensuremath{\pm}0.0001} & \textbf{0.0130\ensuremath{\pm}0.0004} & 0.0081\ensuremath{\pm}0.0004 \\
\midrule
\multirow{5}{*}{Positive forward KL}
 & \(k=1\)  & \textbf{0.822\ensuremath{\pm}0.004} & \underline{0.560\ensuremath{\pm}0.009} & 0.521\ensuremath{\pm}0.019 \\
 & \(k=5\)  & \textbf{0.822\ensuremath{\pm}0.005} & \underline{0.551\ensuremath{\pm}0.004} & 0.513\ensuremath{\pm}0.014 \\
 & \(k=10\) & \textbf{0.831\ensuremath{\pm}0.005} & \underline{0.571\ensuremath{\pm}0.003} & 0.530\ensuremath{\pm}0.015 \\
 & \(k=20\) & \textbf{0.839\ensuremath{\pm}0.005} & \underline{0.560\ensuremath{\pm}0.003} & 0.516\ensuremath{\pm}0.016 \\
 & Random-pair & \underline{0.0355\ensuremath{\pm}0.0001} & \textbf{0.0362\ensuremath{\pm}0.0002} & 0.0354\ensuremath{\pm}0.0010 \\
\midrule
\multirow{5}{*}{Positive reverse-KL}
 & \(k=1\)  & \underline{0.636\ensuremath{\pm}0.008} & \textbf{0.689\ensuremath{\pm}0.007} & 0.479\ensuremath{\pm}0.022 \\
 & \(k=5\)  & \underline{0.456\ensuremath{\pm}0.013} & \textbf{0.639\ensuremath{\pm}0.007} & 0.374\ensuremath{\pm}0.020 \\
 & \(k=10\) & \underline{0.440\ensuremath{\pm}0.013} & \textbf{0.635\ensuremath{\pm}0.007} & 0.363\ensuremath{\pm}0.019 \\
 & \(k=20\) & \underline{0.429\ensuremath{\pm}0.013} & \textbf{0.633\ensuremath{\pm}0.007} & 0.357\ensuremath{\pm}0.019 \\
 & Random-pair & \underline{0.0057\ensuremath{\pm}0.0003} & \textbf{0.0098\ensuremath{\pm}0.0004} & 0.0057\ensuremath{\pm}0.0003 \\
\bottomrule
\end{tabular}
\caption{Recurrence between steps 20 and 100 as a function of the number of selected vocabulary entries. Values are means $\pm$ sample standard deviations over 3 training seeds. Random-pair rows use $k=10$ with mismatched problems.}
\label{tab:recurrence-sensitivity}
\end{table*}

\paragraph{Reinforcement learning.}
Reinforcement learning trains a policy using feedback on its sampled responses. Unlike distillation, it relies on interaction with an environment to obtain training signals. PPO and GRPO are two representative methods.
PPO updates the policy with a clipped advantage objective~\citep{schulman2017ppo}. GRPO computes advantages relative to groups of sampled responses without an explicit critic model~\citep{shao2024deepseekmath}.
Because rewards for large language models are often sparse, several methods combine reinforcement learning with distillation or fine-tuning to provide additional supervision.
AOPD keeps policy gradient updates on tokens with positive teacher advantage and replaces them with truncated forward KL minimization toward the teacher on the remaining tokens~\citep{jia2026aopd}. Uni-OPD trains the student with teacher guidance calibrated by response outcomes~\citep{hou2026uniopd}. REOPOLD trains the student using clipped teacher rewards and selectively sampled tokens~\citep{ko2026reopold}. \citet{akhondzadeh2026rewardgated} apply teacher distillation to a trajectory only when its verifier reward and the teacher--student likelihood gap agree in direction.

Rewards can also be used to construct distillation targets or dense training signals.
BOND trains the student to match a distribution induced by Best-of-$N$ selection~\citep{sessa2025bond}. Direct-OPD uses the log-probability ratio of a weak model before and after reinforcement learning as a dense reward on the student's own trajectories~\citep{feng2026directopd}.

\begin{table*}[t]

\centering
\small
\setlength{\tabcolsep}{3pt}
\begin{tabular}{llrrr}
\toprule
Method & Step & Clip rate & $D_{\mathrm{KL}}(p_T\|p_S)$ & $D_{\mathrm{KL}}(p_S\|p_T)$ \\
\midrule
\multirow{5}{*}{OPSD}
& 20 & \underline{0.565\ensuremath{\pm}0.002} & \textbf{0.192\ensuremath{\pm}0.001} & \textbf{0.110\ensuremath{\pm}0.002} \\
& 40 & \textbf{0.545\ensuremath{\pm}0.003} & \textbf{1.626\ensuremath{\pm}0.040} & \textbf{0.311\ensuremath{\pm}0.002} \\
& 60 & \textbf{0.558\ensuremath{\pm}0.007} & \textbf{2.132\ensuremath{\pm}0.133} & \textbf{0.300\ensuremath{\pm}0.014} \\
& 80 & \underline{0.569\ensuremath{\pm}0.006} & \textbf{2.053\ensuremath{\pm}0.125} & \textbf{0.273\ensuremath{\pm}0.013} \\
& 100 & 0.569\ensuremath{\pm}0.001 & \textbf{2.042\ensuremath{\pm}0.058} & \textbf{0.261\ensuremath{\pm}0.003} \\
\midrule
\multirow{5}{*}{\methodplain{}}
& 20 & \textbf{0.663\ensuremath{\pm}0.004} & \underline{0.152\ensuremath{\pm}0.006} & \underline{0.071\ensuremath{\pm}0.003} \\
& 40 & \underline{0.510\ensuremath{\pm}0.004} & \underline{0.655\ensuremath{\pm}0.009} & 0.137\ensuremath{\pm}0.000 \\
& 60 & \underline{0.553\ensuremath{\pm}0.007} & \underline{0.767\ensuremath{\pm}0.079} & \underline{0.124\ensuremath{\pm}0.011} \\
& 80 & \textbf{0.612\ensuremath{\pm}0.090} & \underline{0.711\ensuremath{\pm}0.092} & \underline{0.114\ensuremath{\pm}0.008} \\
& 100 & \textbf{0.613\ensuremath{\pm}0.026} & 0.641\ensuremath{\pm}0.116 & 0.106\ensuremath{\pm}0.018 \\
\midrule
\multirow{5}{*}{\method{}}
& 20 & 0.517\ensuremath{\pm}0.005 & 0.094\ensuremath{\pm}0.003 & 0.064\ensuremath{\pm}0.002 \\
& 40 & 0.309\ensuremath{\pm}0.002 & 0.375\ensuremath{\pm}0.001 & \underline{0.149\ensuremath{\pm}0.001} \\
& 60 & 0.469\ensuremath{\pm}0.023 & 0.669\ensuremath{\pm}0.046 & 0.112\ensuremath{\pm}0.005 \\
& 80 & 0.512\ensuremath{\pm}0.059 & 0.674\ensuremath{\pm}0.071 & 0.109\ensuremath{\pm}0.002 \\
& 100 & \underline{0.608\ensuremath{\pm}0.093} & \underline{0.764\ensuremath{\pm}0.163} & \underline{0.122\ensuremath{\pm}0.024} \\
\bottomrule
\end{tabular}
\caption{Checkpoint diagnostics, reported as means $\pm$ sample standard deviations over 3 training seeds. Clip rate is the fraction of positive forward KL contribution mass removed by the student cap. The KL columns report full-vocabulary divergences. At each checkpoint, the highest value for each metric is bold and the second highest is underlined.}
\label{tab:diagnostic-ci}
\end{table*}

\paragraph{Teacher construction and privileged guidance.}
Teacher construction varies in how the teacher is trained and which information it observes. Some methods update the teacher together with the student.
\citet{anil2018codistillation} train peer models to match one another's predictions during parallel training. $\pi$-Distill jointly trains a privileged teacher and an unprivileged student with shared parameters~\citep{penaloza2026pidistill}. PAST trains the student against a teacher adapted using completed student trajectories and verified outcomes~\citep{feng2026past}. VISTA trains the student against a privileged teacher that is adapted toward the student on verified rollouts at positions with large KL divergence~\citep{ding2026vista}.

Other methods alter the teacher's context or select among sources of guidance.
ATESD trains the student against a teacher whose exposure to the verified solution is adjusted during training~\citep{han2026atesd}. MOPD trains the student using teacher predictions conditioned on peer attempts~\citep{yu2026mopd}. CANON minimizes the divergence between the student and a teacher conditioned on a consensus answer~\citep{gkountouras2026canon}. DOPD trains the student with token-wise guidance selected from either a privileged teacher or a privileged student~\citep{yu2026dopd}.

Privileged residuals can also be used to construct training targets.
AR-OPD trains the student against a target that combines a partially privileged anchor with an oracle residual~\citep{zhang2026aropd}. AVSD trains the student using residual guidance from privileged views that agree with a shared signal~\citep{nguyen2026avsd}.

\section{OPSD mismatch diagnostics}
\label{app:diagnostics}

We measure distributional mismatch, temporal recurrence, and clipping on Qwen3-4B checkpoints at steps 20, 40, 60, 80, and 100. The resulting diagnostics are summarized in Figure~\ref{fig:diagnostic}.

\subsection{KL and recurrence protocol}
For each checkpoint in the full KL curves, we average over 160 samples from the final five rollout batches in the corresponding 20-step window. Recurrence is measured on 500 fixed response texts with the Qwen3-4B threshold $\tau=10^{-6}$. At token position $i$, the forward and reverse KL contributions for vocabulary entry $v$ are
\begin{equation}
c^{\mathrm{f}}_{i,v}
=p_{T,i}(v)\log\frac{p_{T,i}(v)}{p_{S,i}(v)}.
\end{equation}
\begin{equation}
c^{\mathrm{r}}_{i,v}
=p_{S,i}(v)\log\frac{p_{S,i}(v)}{p_{T,i}(v)}.
\end{equation}

For clipped forward recurrence, $C_{i,t}^{(k)}$ contains the top $k$ entries ranked by $c^{\mathrm{f}}_{i,v}$ that also satisfy $c^{\mathrm{f}}_{i,v}>\tau$. Positive forward and reverse recurrence rank entries by the corresponding contribution and use cutoffs $c^{\mathrm{f}}_{i,v}>0$ and $c^{\mathrm{r}}_{i,v}>0$. At checkpoints $a$ and $b$, recurrence is the mean Jaccard overlap over the $N$ valid token positions,
\begin{equation}
R_k(a,b)=\frac{1}{N}\sum_{i=1}^{N}
\frac{|C_{i,a}^{(k)}\cap C_{i,b}^{(k)}|}
{|C_{i,a}^{(k)}\cup C_{i,b}^{(k)}|}.
\end{equation}

The primary result in Table~\ref{tab:recurrence-sensitivity} uses $a=20$, $b=100$, and $k=10$. Each random-pair baseline uses a deterministic derangement to pair every step-20 problem with a different step-100 problem, preserving the marginal vocabulary distributions for comparison with recurrence on the same problems.

The results for the same problems are stable across $k$. Clipped forward recurrence ranges from 0.682 to 0.720 for OPSD, from 0.338 to 0.445 for \methodplain{}, and from 0.182 to 0.314 for \method{}. Positive forward recurrence follows the same ordering: OPSD ranges from 0.822 to 0.839, and both adaptive variants have lower recurrence. For positive reverse KL contributions, \methodplain{} is highest across all four values of $k$, and \method{} is lowest. For every signal and method at $k=10$, recurrence on the same problems substantially exceeds the random-pair baseline.

\begin{figure*}[t]
    \centering
    \includegraphics[width=\textwidth]{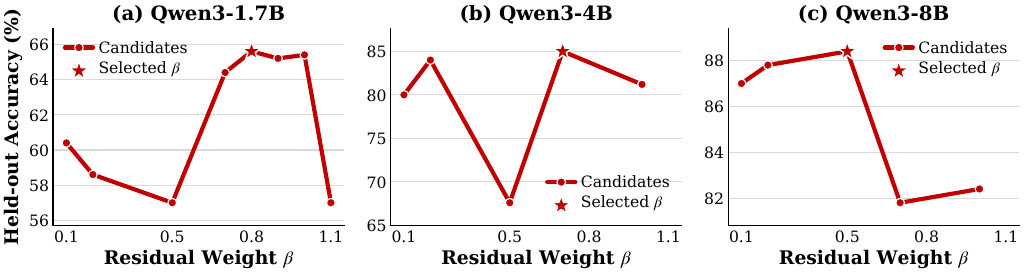}
    \caption{Held-out validation accuracy as a function of residual weight at each Qwen3 scale. Each point represents the step-100 checkpoint trained with seed 42, evaluated using one judged response per problem on the 500-problem validation set. Stars mark the weights selected by the rule in Table~\ref{tab:beta-validation}.}
    \label{fig:beta-validation}
\end{figure*}

\subsection{Clipping and divergence details}
Let $c^+_{i,v}=\max(c^{\mathrm{f}}_{i,v},0)$. The student clip rate at token position $i$ is the fraction of positive contribution mass removed by the cap,
\begin{equation}
\rho_i=1-\frac{\sum_v\min(c^+_{i,v},\tau)}{\sum_v c^+_{i,v}},
\end{equation}
excluding positions with zero positive mass.

Table~\ref{tab:diagnostic-ci} gives the checkpoint values summarized in Figure~\ref{fig:diagnostic}: the mean of $\rho_i$ and the full-vocabulary forward and reverse KL divergences. Both adaptive teacher variants have lower forward and reverse KL than OPSD at step 100.

Under OPSD, forward KL increases from 0.192 at step 20 to 2.042 at step 100, and reverse KL increases from 0.110 to 0.261. At step 100, \methodplain{} reduces these divergences to 0.641 and 0.106, and \method{} gives 0.764 and 0.122, respectively. Both adaptive teacher variants maintain substantially lower divergence than OPSD in both directions at the later checkpoints. 

\section{Residual weight selection and sensitivity}
\label{app:beta-selection}

The residual weight $\beta$ controls how much of the frozen privileged reference effect is retained in the teacher target. Equation~\ref{eq:residual} defines the frozen initial model's log-probability residual $\Delta_n^0$. Equations~\ref{eq:target-score} and~\ref{eq:target} add $\beta\Delta_n^0$ to the updated student's log probabilities and normalize the result. Setting $\beta=0$ recovers \methodplain{}.

\subsection{Held-out selection}

\begin{table}[h]
\centering
\small
\setlength{\tabcolsep}{4pt}
\begin{tabular}{llrrrr}
\toprule
Model & $\beta$ & Corr. & Acc. (\%) & Trunc. (\%) & Tokens \\
\midrule
\multirow{8}{*}{1.7B}
& 0.1 & 302 & 60.4 & 1.6 & 1,394 \\
& 0.2 & 293 & 58.6 & 3.0 & 1,414 \\
& 0.5 & 285 & 57.0 & 4.0 & 1,359 \\
& 0.7 & 322 & 64.4 & 15.6 & 3,053 \\
& \textbf{0.8} & \textbf{328} & \textbf{65.6} & 6.0 & 1,834 \\
& 0.9 & 326 & 65.2 & 5.4 & 1,696 \\
& 1.0 & 327 & 65.4 & 6.6 & 2,149 \\
& 1.1 & 285 & 57.0 & 9.4 & 2,023 \\
\midrule
\multirow{5}{*}{4B}
& 0.1 & 400 & 80.0 & 1.4 & 1,020 \\
& 0.2 & 420 & 84.0 & 2.2 & 1,575 \\
& 0.5 & 338 & 67.6 & 0.4 & 627 \\
& \textbf{0.7} & \textbf{425} & \textbf{85.0} & 1.4 & 1,219 \\
& 1.0 & 406 & 81.2 & 3.6 & 2,002 \\
\midrule
\multirow{5}{*}{8B}
& 0.1 & 435 & 87.0 & 3.4 & 1,508 \\
& 0.2 & 439 & 87.8 & 1.4 & 1,229 \\
& \textbf{0.5} & \textbf{442} & \textbf{88.4} & 2.8 & 1,723 \\
& 0.7 & 409 & 81.8 & 6.0 & 1,760 \\
& 1.0 & 412 & 82.4 & 2.2 & 1,028 \\
\bottomrule
\end{tabular}
\caption{Residual weight search on the held-out validation set for three Qwen3 models. Corr.\ counts correct responses out of 500, and Tokens is the mean observed response length. Bold entries mark the selected settings.}
\label{tab:beta-validation}
\end{table}

We select residual weights for the 1.7B, 4B, and 8B models on a fixed validation set of 500 OpenThoughts problems held out from training. This set is disjoint from training rollouts and from AIME24, AIME25, and HMMT25. The initial search uses $\beta\in\{0.1,0.2,0.5,0.7,1.0\}$ at each scale. For 1.7B, we further add $\beta\in\{0.8,0.9,1.1\}$. 

For each candidate, we evaluate the step-100 checkpoint trained with seed 42 using one judged response per validation problem. We select the candidate with the highest judged accuracy, yielding weights of $0.8$, $0.7$, and $0.5$, respectively. Table~\ref{tab:beta-validation} lists all candidates.

\begin{table}[h]
\centering
\small
\setlength{\tabcolsep}{4pt}
\begin{tabular}{lccc}
\toprule
$\beta$ & avg@12 / pass@12 & Trunc. (\%) & Tokens \\
\midrule
0 & 59.44 / 83.33 & 3.89 & 6.80K \\
0.1 & 55.56 / 80.00 & 5.28 & 5.68K \\
0.2 & 51.11 / 80.00 & 5.56 & 6.21K \\
0.5 & 60.56 / 83.33 & 5.28 & 7.33K \\
0.7 & 45.00 / 80.00 & 15.93 & 6.01K \\
1.0 & 34.17 / 66.67 & 18.61 & 3.85K \\
\bottomrule
\end{tabular}
\caption{Qwen3-8B residual weight sensitivity on AIME 2024. $\beta=0$ corresponds to \methodplain{}. Tokens is the mean length of non-truncated responses. All rows use training seed 42.}
\label{tab:beta}
\end{table}

Figure~\ref{fig:beta-validation} shows that held-out validation accuracy varies nonmonotonically with $\beta$ at all three scales. For 1.7B, accuracy rises from 57.0\% at $\beta=0.5$ to 65.6\% at the selected $\beta=0.8$, while the neighboring settings $0.9$ and $1.0$ remain close at 65.2\% and 65.4\%. For 4B, the selected $\beta=0.7$ reaches 85.0\%, compared with 84.0\% at $0.2$ and 67.6\% at $0.5$. For 8B, the selected $\beta=0.5$ reaches 88.4\%, compared with 87.8\% at $0.2$, before accuracy falls to 81.8\% and 82.4\% at $0.7$ and $1.0$. In these sweeps, the weight with the highest validation accuracy decreases from $0.8$ to $0.7$ to $0.5$ as model scale increases.

\subsection{Sensitivity on the test set}

Table~\ref{tab:beta} reports a separate Qwen3-8B sensitivity analysis on AIME 2024. Here, avg@12 is the fraction of all $30\times12$ responses that are correct, and pass@12 is the fraction of test problems with at least one correct response. The mean token count excludes truncated responses. At $\beta=0.5$, avg@12 exceeds the result at $\beta=1$ by 26.39 percentage points (60.56\% versus 34.17\%). The choice of $\beta=0.5$ comes from the held-out validation results above.

\section{Reproducibility}
\label{app:repro}

This section specifies the prompts, optimization schedule, and evaluation rules used for every result in the paper. Unless stated otherwise, results of trained methods use three training seeds, and benchmark metrics use 12 sampled responses per problem.

\subsection{Exact prompts}
The student user message is:
\begin{quote}\small\ttfamily
Problem: \{problem\}\par
\medskip
Please reason step by step, and put your final answer within \textbackslash boxed\{\}.
\end{quote}
The teacher user message is:
\begin{quote}\small\ttfamily
Problem: \{problem\}\par
\medskip
Here is a reference solution to this problem:\par
=== Reference Solution Begin ===\par
\{reference\}\par
=== Reference Solution End ===\par
\medskip
After reading the reference solution above, make sure you truly understand the reasoning behind each step --- do not copy or paraphrase it. Now, using your own words and independent reasoning, derive the same final answer to the problem above. Think step by step, explore different approaches, and don't be afraid to backtrack or reconsider if something doesn't work out:\par
\medskip
Please reason step by step, and put your final answer within \textbackslash boxed\{\}.
\end{quote}
Both messages use the Qwen chat template with a generation prompt appended and thinking mode disabled. Benchmark evaluation uses the same final-answer instruction.

\subsection{Implementation details}

\begin{table*}[t]
\centering
\small
\setlength{\tabcolsep}{3.2pt}
\begin{tabular}{lrrrr}
\toprule
Method & AIME24 & AIME25 & HMMT25 & Mean \\
\midrule
\multicolumn{5}{l}{\textit{Qwen3-1.7B}} \\
Base & 11.11 / 14.17 / 10.56 & 8.33 / 10.56 / 9.72 & 5.83 / 5.28 / 6.11 & 8.43 / 10.00 / 8.80 \\
SFT & 9.44 / 7.50 / 9.44 & 7.50 / 7.50 / 8.61 & 2.22 / 1.11 / 1.67 & 6.39 / 5.37 / 6.57 \\
GRPO & 13.06 / 9.44 / 15.00 & 10.28 / 10.28 / 8.61 & 5.00 / 4.44 / 5.00 & 9.44 / 8.06 / 9.54 \\
OPSD & 15.56 / 14.17 / 12.78 & 10.00 / 8.89 / 10.28 & 6.39 / 7.78 / 5.28 & 10.65 / 10.28 / 9.44 \\
OPSD, matched & 11.67 / 10.56 / 16.94 & 7.22 / 7.50 / 8.33 & 6.11 / 7.22 / 5.83 & 8.33 / 8.43 / 10.37 \\
\methodplain{} & 11.94 / 12.50 / 10.28 & 8.06 / 9.72 / 7.78 & 4.44 / 5.56 / 4.17 & 8.15 / 9.26 / 7.41 \\
\method{} & 13.33 / 13.61 / 18.06 & 11.67 / 10.83 / 14.17 & 9.17 / 5.28 / 8.33 & 11.39 / 9.91 / 13.52 \\
\midrule
\multicolumn{5}{l}{\textit{Qwen3-4B}} \\
Base & 22.78 / 24.72 / 20.28 & 20.83 / 21.94 / 17.50 & 10.28 / 10.83 / 8.89 & 17.96 / 19.17 / 15.56 \\
SFT & 19.44 / 18.33 / 21.67 & 18.06 / 16.94 / 19.44 & 10.83 / 10.28 / 8.06 & 16.11 / 15.19 / 16.39 \\
GRPO & 21.39 / 23.06 / 22.50 & 18.61 / 16.94 / 20.00 & 10.28 / 12.50 / 11.11 & 16.76 / 17.50 / 17.87 \\
OPSD & 28.61 / 28.06 / 29.44 & 26.11 / 24.17 / 24.17 & 15.56 / 13.33 / 13.06 & 23.43 / 21.85 / 22.22 \\
OPSD, matched & 26.67 / 26.94 / 31.11 & 21.11 / 21.67 / 24.72 & 14.72 / 12.78 / 13.89 & 20.83 / 20.46 / 23.24 \\
\methodplain{} & 41.67 / 43.89 / 31.67 & 31.11 / 30.83 / 28.89 & 22.22 / 21.94 / 15.83 & 31.67 / 32.22 / 25.46 \\
\method{} & 43.33 / 41.39 / 43.61 & 33.06 / 41.39 / 37.78 & 21.94 / 25.83 / 25.00 & 32.78 / 36.20 / 35.46 \\
\midrule
\multicolumn{5}{l}{\textit{Qwen3-8B}} \\
Base & 27.50 / 25.00 / 26.67 & 19.17 / 19.72 / 18.89 & 11.67 / 10.28 / 11.11 & 19.44 / 18.33 / 18.89 \\
SFT & 22.22 / 23.89 / 20.00 & 17.78 / 16.39 / 15.83 & 8.33 / 7.78 / 9.17 & 16.11 / 16.02 / 15.00 \\
GRPO & 26.94 / 29.44 / 26.11 & 18.33 / 19.17 / 20.56 & 12.22 / 10.56 / 10.56 & 19.17 / 19.72 / 19.07 \\
OPSD & 35.83 / 40.56 / 35.28 & 27.78 / 28.89 / 25.56 & 16.67 / 14.72 / 14.72 & 26.76 / 28.06 / 25.19 \\
OPSD, matched & 36.39 / 31.94 / 33.06 & 24.44 / 24.44 / 25.28 & 13.33 / 11.94 / 11.94 & 24.72 / 22.78 / 23.43 \\
\methodplain{} & 59.44 / 56.67 / 55.83 & 41.67 / 38.06 / 39.72 & 26.67 / 26.94 / 26.11 & 42.59 / 40.56 / 40.56 \\
\method{} & 60.56 / 54.72 / 59.44 & 43.89 / 40.28 / 45.00 & 26.94 / 20.83 / 29.72 & 43.80 / 38.61 / 44.72 \\
\bottomrule
\end{tabular}
\caption{Qwen3-1.7B, 4B, and 8B avg@12 results by seed (\%). Values are ordered by seed 42 / 43 / 44.}
\label{tab:seed-results}
\end{table*}

\paragraph{Shared training configuration.}
All trained methods use the same OpenThoughts split and training seeds 42, 43, and 44. Each run uses one NVIDIA A100 80GB GPU. The data seed is fixed to 42. We train in bfloat16 with gradient checkpointing and optimize LoRA parameters only. Each LoRA adapter has rank 64 and scale 128 and is applied to the query, key, value, output, gate, up, and down projections. LoRA dropout is 0.05 for SFT and GRPO and 0 for the distillation methods. We use the AdamW optimizer with a learning rate of $5\times10^{-6}$. For the distillation methods, we use 10 warmup steps followed by a constant learning rate, zero weight decay, and a maximum gradient norm of 0.1.

\subsubsection{Baseline methods}

\paragraph{Base and SFT.}
Base directly evaluates the pretrained checkpoint. SFT uses the problem as the user input and the verified reference solution as the assistant target. We train SFT for 100 optimizer steps with a global batch size of 64 and a maximum sequence length of 16,000 tokens.

\paragraph{GRPO.}
GRPO assigns a binary reward by extracting the boxed final answer. It is trained for 200 optimizer steps with an effective batch size of 32. For every prompt, it samples eight completions at temperature 1.0 and performs two update iterations. We use the GRPO loss with group normalization and no reference KL penalty. We use a prompt limit of 2,048 tokens and a completion limit of 8,192 tokens.

\paragraph{OPSD.}
OPSD is trained for 100 steps. Each step uses 32 prompts and one completion per prompt from the current student policy. Rollouts use temperature 1.0, top-$p$ 0.95, top-$k$ 20, and a completion limit of 1,024 tokens. The frozen privileged teacher scores the same completion under the reference-conditioned prompt. We optimize the student using the forward KL in Equation~\ref{eq:student}. The pointwise upper clipping threshold $\tau$ is $10^{-6}$ for Qwen3-1.7B and Qwen3-4B and $10^{-7}$ for Qwen3-8B.

\subsubsection{Our methods}

The student and teacher use independently initialized LoRA adapters on one frozen backbone, with separate optimizers. The rollout and student update are identical to OPSD, including its sampling configuration and pointwise student clip. Student and teacher KL losses are computed over the full vocabulary at every valid completion position. Token losses are normalized within groups of four responses and then averaged over the eight groups in the effective batch of 32.

After the student update, we recompute the student logits on the same sampled token IDs without gradients, then update the teacher. In \methodplain{}, the privileged teacher minimizes the unclipped $D_{\mathrm{KL}}(p_S\|p_T)$ objective in Equation~\ref{eq:teacher-direct}. In \method{}, we also score the frozen backbone with and without the reference to form $\Delta^0$. The teacher then minimizes the unclipped full-vocabulary KL from the stopped residual target in Equation~\ref{eq:teacher}. The student, the two frozen backbone branches, and the constructed target are detached during the teacher update, so gradients are applied only to the current teacher adapter. Only the student adapter is merged with the backbone for evaluation.

\begin{table*}[t]

\centering
\small
\setlength{\tabcolsep}{2.5pt}
\begin{tabular}{llrrrrrr}
\toprule
Model & Method & PFLOPs & GPU h & $\Delta$ Acc. & $\Delta$ tokens & Trunc. (\%)$\downarrow$ & $E\uparrow$ \\
\midrule
\multirow{4}{*}{1.7B}
& OPSD & $66.62$ & 0.61 & $+1.05$ & $+1.94\mathrm{K}$ & $2.90$ & 0.54 \\
& OPSD, matched compute & $181.35$ & 2.20 & $-0.03$ & $+3.58\mathrm{K}$ & $9.07$ & -0.01 \\
& \methodplain{} & $128.61$ & 0.81 & $-0.80$ & $+0.67\mathrm{K}$ & $1.05$ & -1.21 \\
& \method{} & $180.94$ & 1.05 & $+2.53$ & $+5.32\mathrm{K}$ & $8.77$ & 0.48 \\
\midrule
\multirow{4}{*}{4B}
& OPSD & $146.28$ & 1.01 & $+4.94$ & $+4.10\mathrm{K}$ & $8.49$ & 1.20 \\
& OPSD, matched compute & $395.12$ & 2.83 & $+3.95$ & $+6.20\mathrm{K}$ & $18.80$ & 0.64 \\
& \methodplain{} & $281.41$ & 1.40 & $+12.22$ & $+2.73\mathrm{K}$ & $2.35$ & 4.47 \\
& \method{} & $394.46$ & 1.78 & $+17.25$ & $+5.28\mathrm{K}$ & $5.68$ & 3.27 \\
\midrule
\multirow{4}{*}{8B}
& OPSD & $272.43$ & 1.38 & $+7.78$ & $+3.76\mathrm{K}$ & $7.96$ & 2.07 \\
& OPSD, matched compute & $733.10$ & 4.53 & $+4.75$ & $+4.21\mathrm{K}$ & $11.54$ & 1.13 \\
& \methodplain{} & $545.74$ & 2.00 & $+22.35$ & $+6.00\mathrm{K}$ & $3.86$ & 3.72 \\
& \method{} & $732.02$ & 2.52 & $+23.49$ & $+5.94\mathrm{K}$ & $5.59$ & 3.95 \\
\bottomrule
\end{tabular}
\caption{Per-scale cost and generation behavior for distillation methods. $\Delta$ Acc. is the change in the avg@12 mean relative to the corresponding Base mean in percentage points. $\Delta$ tokens is the change in mean observed completion tokens relative to that Base mean. Length efficiency $E$ is defined in Equation~\ref{eq:length-efficiency}.}
\label{tab:tradeoff-by-scale}
\end{table*}

\subsection{Evaluation metrics}
Let $z_{b,i,j}\in\{0,1\}$ indicate whether response $j$ to problem $i$ on benchmark $b$ is correct. With $P=30$ problems and $J=12$ responses per problem, we compute
\begin{equation}
\operatorname{avg@12}_b
=\frac{1}{PJ}\sum_{i=1}^{P}\sum_{j=1}^{J}z_{b,i,j}.
\end{equation}
\begin{equation}
\operatorname{pass@12}_b
=\frac{1}{P}\sum_{i=1}^{P}
\mathbf{1}\!\left[\sum_{j=1}^{J}z_{b,i,j}>0\right].
\end{equation}
Reported accuracy is the arithmetic mean of avg@12 across AIME24, AIME25, and HMMT25. For a statistic $u_s$ measured on training seeds $s\in\{42,43,44\}$, tables report $\bar u\pm s_u$, where $\bar u=\frac{1}{3}\sum_su_s$ and $s_u=\sqrt{\frac{1}{2}\sum_s(u_s-\bar u)^2}$ is the sample standard deviation.

For each problem, the total context limit is 32,768 tokens, including the prompt. The completion limit is the remaining budget after prompt tokenization. A response is truncated when its finish reason is \texttt{length} or it reaches that remaining budget. We extract the last balanced \texttt{\textbackslash boxed\{\ldots\}} expression and parse it with \texttt{math\_verify}.

\section{Results by training seed}
\label{app:seed}

Table~\ref{tab:seed-results} reports the results of each seed underlying Table~\ref{tab:main}. For Base, the seeds are generation seeds. The Mean column is the arithmetic mean across AIME24, AIME25, and HMMT25 within each seed. ``OPSD, matched'' uses the checkpoint at each scale whose PFLOPs are closest to those of the step-100 \method{} run.

\section{Compute and generation details}
\label{app:cost}

This section provides the cost and generation statistics by model scale underlying the aggregate results in Table~\ref{tab:tradeoff}.

\subsection{Cost measurement}

We measure training cost using elapsed wall time and estimated PFLOPs.
Wall time includes rollout generation, scoring, optimization, checkpoint I/O, and systems overhead. Let $N$ be the backbone parameter count and $T_\ell$ the number of tokens processed by model call $\ell$. We estimate compute as $\sum_\ell a_\ell NT_\ell/10^{15}$ PFLOPs, with $a_\ell=2$ for a frozen forward pass and $a_\ell=6$ for a trainable forward and backward pass.

\begin{table}[h]
\centering
\small
\setlength{\tabcolsep}{3pt}
\begin{tabular}{llrr}
\toprule
Model & Method & GPU h & PFLOPs \\
\midrule
\multirow{3}{*}{1.7B} & OPSD & 0.61 & $66.62{\pm}0.32$ \\
& \methodplain{} & 0.81 & $128.61{\pm}3.75$ \\
& \method{} & 1.05 & $180.94{\pm}0.08$ \\
\midrule
\multirow{3}{*}{4B} & OPSD & 1.01 & $146.28{\pm}1.06$ \\
& \methodplain{} & 1.40 & $281.41{\pm}8.23$ \\
& \method{} & 1.78 & $394.46{\pm}9.08$ \\
\midrule
\multirow{3}{*}{8B} & OPSD & 1.38 & $272.43{\pm}2.10$ \\
& \methodplain{} & 2.00 & $545.74{\pm}9.64$ \\
& \method{} & 2.52 & $732.02{\pm}11.10$ \\
\bottomrule
\end{tabular}
\caption{Training costs for 100-step distillation runs. GPU hours are measured from elapsed wall time, and PFLOPs are estimated from model calls and processed tokens.}
\label{tab:full-compute}
\end{table}

Table~\ref{tab:full-compute} reports training costs for 100-step distillation runs. GPU hours are elapsed hours multiplied by the number of GPUs.

\begin{table*}[t]

\centering
\small
\begin{tabular}{llrr}
\toprule
Model & Method & Untruncated tokens & Truncation (\%) \\
\midrule
\multirow{3}{*}{1.7B} & OPSD & $3.41\mathrm{K}\pm0.24$ & $2.90\pm0.33$ \\
& \methodplain{} & $2.67\mathrm{K}\pm0.55$ & $1.05\pm0.19$ \\
& \method{} & $5.21\mathrm{K}\pm0.96$ & $8.77\pm3.23$ \\
\midrule
\multirow{3}{*}{4B} & OPSD & $4.72\mathrm{K}\pm0.26$ & $8.49\pm2.73$ \\
& \methodplain{} & $5.06\mathrm{K}\pm0.95$ & $2.35\pm0.88$ \\
& \method{} & $6.82\mathrm{K}\pm1.16$ & $5.68\pm2.52$ \\
\midrule
\multirow{3}{*}{8B} & OPSD & $4.34\mathrm{K}\pm0.30$ & $7.96\pm1.49$ \\
& \methodplain{} & $7.89\mathrm{K}\pm0.16$ & $3.86\pm0.56$ \\
& \method{} & $7.39\mathrm{K}\pm1.04$ & $5.59\pm2.53$ \\
\bottomrule
\end{tabular}
\caption{Generation length and truncation computed over all responses on the three benchmarks. Table~\ref{tab:tradeoff-by-scale} reports observed token changes including truncated responses.}
\label{tab:full-generation}
\end{table*}

\begin{table*}[t]
\centering
\small
\begin{tabular}{llrrrrrr}
\toprule
Model & Teacher clipping & AIME24 & AIME25 & HMMT25 & Mean & Length (K) & Trunc. (\%) \\
\midrule
\multirow{3}{*}{Qwen3-1.7B}
& High clipping & 11.94 & 6.48 & 4.35 & $7.59\pm0.48$ & $2.13\pm0.16$ & $1.23\pm0.33$ \\
& Low clipping & 13.52 & 10.93 & 5.19 & $9.88\pm0.81$ & $4.40\pm0.67$ & $6.42\pm0.63$ \\
& No clipping & 15.00 & 12.22 & 7.59 & $\mathbf{11.60\pm1.82}$ & $5.22\pm0.96$ & $8.77\pm3.23$ \\
\midrule
\multirow{3}{*}{Qwen3-4B}
& High clipping & 28.70 & 24.81 & 12.96 & $22.16\pm1.72$ & $3.16\pm0.67$ & $5.96\pm1.68$ \\
& Low clipping & 35.74 & 30.56 & 19.07 & $28.46\pm4.88$ & $5.83\pm1.64$ & $8.09\pm4.15$ \\
& No clipping & 42.78 & 37.41 & 24.26 & $\mathbf{34.81\pm1.80}$ & $6.81\pm1.16$ & $5.68\pm2.52$ \\
\midrule
\multirow{3}{*}{Qwen3-8B}
& High clipping & 35.19 & 24.81 & 11.02 & $23.67\pm2.94$ & $2.38\pm0.31$ & $4.10\pm0.84$ \\
& Low clipping & 59.91 & 42.41 & 25.46 & $\mathbf{42.59\pm2.94}$ & $8.12\pm0.40$ & $4.51\pm2.59$ \\
& No clipping & 58.24 & 43.06 & 25.83 & $42.38\pm3.29$ & $7.39\pm1.05$ & $5.59\pm2.53$ \\
\bottomrule
\end{tabular}
\caption{Teacher clipping ablation on the \method{} teacher objective at each scale after 100 training steps. Benchmark entries are avg@12 percentages averaged over training seeds 42, 43, and 44. Mean is the macro average across AIME24, AIME25, and HMMT25. Length is the mean completion length of non-truncated responses. Trunc. is the truncation rate. Mean, Length, and Trunc. report means $\pm$ sample standard deviations across training seeds.}
\label{tab:teacher-clip-scales}
\end{table*}

The additional teacher-side operations increase compute at every model scale. Relative to OPSD, \methodplain{} uses 1.93$\times$, 1.92$\times$, and 2.00$\times$ as many PFLOPs at 1.7B, 4B, and 8B, while \method{} uses 2.72$\times$, 2.70$\times$, and 2.69$\times$. The corresponding ratios of GPU hours are smaller: 1.33$\times$--1.45$\times$ for \methodplain{} and 1.72$\times$--1.83$\times$ for \method{}.

\subsection{Tradeoffs by model scale}

We compare accuracy gains, response lengths, and truncation rates at each model scale. For a method $m$ at a fixed model scale, $\Delta\operatorname{Acc}(m)=\operatorname{Acc}(m)-\operatorname{Acc}(\mathrm{Base})$ and $\Delta\operatorname{Tok}(m)=\operatorname{Tok}(m)-\operatorname{Tok}(\mathrm{Base})$. Here $\operatorname{Acc}$ is the mean avg@12 across the three benchmarks, and $\operatorname{Tok}$ is the mean number of generated tokens over all responses, including truncated responses. The length efficiency is $E(m)=\Delta\operatorname{Acc}(m)/(\Delta\operatorname{Tok}(m)/1000)$. For each model and training seed, we select the OPSD checkpoint whose cumulative training PFLOPs, including all scoring and update passes, are closest to those of the corresponding \method{} run at step 100.

In Table~\ref{tab:tradeoff-by-scale}, the matched OPSD runs have PFLOP estimates close to those of \method{}. In these comparisons, the accuracy gains over Base are larger for \method{} by 2.56, 13.30, and 18.74 percentage points at 1.7B, 4B, and 8B. The matched OPSD runs also use more GPU hours and have higher truncation rates at all three scales. At 4B and 8B, both methods with adaptive teachers have higher length efficiency than both OPSD settings. At 1.7B, \method{} has lower length efficiency than OPSD at step 100 (0.48 versus 0.54) but higher length efficiency than matched compute OPSD (0.48 versus $-0.01$).

\subsection{Completion length and truncation}

Table~\ref{tab:full-generation} reports untruncated completion lengths and truncation rates computed over all responses on the three benchmarks. The observed token mean counts every generated response up to its stopping point, including responses stopped by the length limit. The untruncated mean excludes responses stopped by this limit. The truncation rate is the fraction of all responses that reach the limit.

\section{Teacher clipping ablations}

\label{app:teacher-objectives}

\begin{table*}[t]
\centering
\small
\begin{tabular}{llrrr}
\toprule
Method & Step & Teacher acc. (\%) & Student acc. (\%) & JS retention \\
\midrule
\multirow{5}{*}{\methodplain{}}
& 20  & 93.93\ensuremath{\pm}0.90 & 81.33\ensuremath{\pm}1.62 & 0.579\ensuremath{\pm}0.007 \\
& 40  & 90.73\ensuremath{\pm}0.58 & 82.00\ensuremath{\pm}1.51 & 0.748\ensuremath{\pm}0.007 \\
& 60  & 91.80\ensuremath{\pm}0.53 & 82.47\ensuremath{\pm}2.08 & 0.773\ensuremath{\pm}0.034 \\
& 80  & 85.87\ensuremath{\pm}8.12 & 79.13\ensuremath{\pm}5.13 & 0.842\ensuremath{\pm}0.120 \\
& 100 & 91.07\ensuremath{\pm}2.08 & 84.73\ensuremath{\pm}3.37 & 0.838\ensuremath{\pm}0.092 \\
\midrule
\multirow{5}{*}{\method{}}
& 20  & 94.33\ensuremath{\pm}0.83 & 82.93\ensuremath{\pm}0.83 & 0.689\ensuremath{\pm}0.002 \\
& 40  & 92.20\ensuremath{\pm}0.53 & 83.73\ensuremath{\pm}1.86 & 1.468\ensuremath{\pm}0.004 \\
& 60  & 91.87\ensuremath{\pm}2.66 & 83.40\ensuremath{\pm}4.04 & 1.216\ensuremath{\pm}0.098 \\
& 80  & 92.07\ensuremath{\pm}2.32 & 86.87\ensuremath{\pm}3.01 & 1.377\ensuremath{\pm}0.384 \\
& 100 & 90.67\ensuremath{\pm}1.81 & 86.87\ensuremath{\pm}0.95 & 1.626\ensuremath{\pm}0.358 \\
\bottomrule
\end{tabular}
\caption{Qwen3-4B accuracy and reference sensitivity measurements underlying Figure~\ref{fig:ablation-dynamics}. Accuracy is measured on the fixed diagnostic set of 500 problems. JS retention is averaged over valid rollout problems as defined in the reference sensitivity protocol. Entries are means $\pm$ sample standard deviations across three training seeds.}
\label{tab:role-dynamics-data}
\end{table*}

This section reports the teacher clipping ablation in Table~\ref{tab:core-ablation} separately for each scale and benchmark. All variants use the \method{} teacher objective $D_{\mathrm{KL}}(\sg[q]\|p_T)$ with the residual target and the selected $\beta$ ($0.8$, $0.7$, and $0.5$ for Qwen3-1.7B, 4B, and 8B). Clipping acts on each vocabulary level contribution $q(v)\log[q(v)/p_T(v)]$ before the vocabulary sum. High clipping caps each contribution at the threshold, low clipping floors it at the threshold, and no clipping leaves all contributions unchanged. No clipping is the main \method{} configuration. The teacher thresholds are equal to the student clipping thresholds: $10^{-6}$ for Qwen3-1.7B and 4B and $10^{-7}$ for Qwen3-8B. 

In Table~\ref{tab:teacher-clip-scales}, we found that high clipping gave the lowest mean accuracy and the shortest responses at every scale. No clipping had the highest mean accuracy at Qwen3-1.7B ($11.60$ versus $9.88$ for low clipping) and Qwen3-4B ($34.81$ versus $28.46$). At Qwen3-8B, low clipping and no clipping had close results. Averaged over the three scales, no clipping gave $29.60\pm1.52$, low clipping $26.98\pm1.13$, and high clipping $17.81\pm0.68$. Based on this average, we use no clipping for the teacher update in the main method.

\section{Teacher adaptation diagnostics}

\label{app:teacher-diagnostics}

\subsection{Teacher accuracy and reference sensitivity}

\paragraph{Accuracy.}

On Qwen3-4B, we evaluated teacher and student accuracy on a fixed diagnostic set of 500 problems at steps 20, 40, 60, 80, and 100 (Figure~\ref{fig:ablation-dynamics}). For each method, checkpoint, and training seed, the teacher received the problem and reference, and the student received only the problem. Each generated one response per problem using generation seed 42, temperature 1.0, top-$p$ 0.95, top-$k$ 20, a completion limit of 8,192 tokens, and a total limit of 16,384 tokens. A deterministic Qwen3-8B judge with thinking mode disabled checked each final answer and assigned binary correctness. Accuracy was the fraction of responses judged correct among the 500 problems. Table~\ref{tab:role-dynamics-data} reports means and sample standard deviations over three training seeds.

\paragraph{Reference sensitivity.}
We measured sensitivity to the reference using student trajectories from the final five rollout batches in each 20-step checkpoint window. For distributions $P$ and $Q$, let $M=(P+Q)/2$. Their Jensen--Shannon (JS) divergence is

\begin{equation}
D_{\mathrm{JS}}(P\|Q)=\tfrac{1}{2}D_{\mathrm{KL}}(P\|M)+\tfrac{1}{2}D_{\mathrm{KL}}(Q\|M).
\end{equation}

For a policy $p$, let $p_{n,r}=p(\cdot\mid h_n,r)$ and $p_n=p(\cdot\mid h_n)$ denote the next-token distributions with and without the reference $r$, evaluated at the same rollout history $h_n$. The reference sensitivity at position $n$ is

\begin{equation}
d_n(p)=D_{\mathrm{JS}}(p_{n,r}\|p_n).
\end{equation}

For each valid retokenized rollout problem $j$, we average the divergence over its completion positions $I_j$:

\begin{equation}
J_j(p)=\frac{1}{|I_j|}\sum_{n\in I_j}d_n(p).
\end{equation}

We then normalize the adapted teacher's sensitivity by that of the frozen initial model $p^0$:

\begin{equation}
\operatorname{JSRet}_j(\phi)
=\frac{J_j(p_T^\phi)}{J_j(p^0)}.
\end{equation}

We average $\operatorname{JSRet}_j$ over valid problems within each training seed and report the mean and sample standard deviation across the three seeds. A ratio of one indicates the same reference sensitivity as the initial model. Ratios below or above one indicate attenuation or amplification. JS retention measures the magnitude of the change induced by the reference in the output distribution.

In Table~\ref{tab:role-dynamics-data}, the teacher had higher mean accuracy than the student at every checkpoint for both methods. Averaged over the five checkpoints, teacher and student accuracy were 90.68\% and 81.93\% for \methodplain{}, and 92.23\% and 84.76\% for \method{}. The corresponding teacher--student gaps averaged 8.75 and 7.47 percentage points.

For reference sensitivity, the mean JS retention for \methodplain{} remained below one at all five checkpoints. It exceeded one from step 40 onward and reached 1.626 at step 100 in \method{}. These measurements show that \method{} retained or amplified reference sensitivity after step 20 while maintaining higher teacher than student mean accuracy.

\paragraph{Clipped forward-KL mass.}
The right panel of Figure~\ref{fig:ablation-dynamics} reports the fraction of positive forward-KL mass removed by the student cap. At token position $n$, define the positive vocabulary-level contribution $c^+_{n,v}=\max\{p_{T,n}(v)\log[p_{T,n}(v)/p_{S,n}(v)],0\}$. The removed fraction is

\begin{equation}
\rho_n=1-\frac{\sum_v\min(c^+_{n,v},\tau)}{\sum_v c^+_{n,v}},
\end{equation}

where $\tau$ is the student clipping threshold. We exclude positions with zero positive mass, average $\rho_n$ within each problem, and then average over valid problems within each seed. Table~\ref{tab:role-clipped-mass} reports the mean and sample standard deviation across training seeds. The final row is the unweighted mean of the five checkpoint means and determines the bar height in the figure.

\begin{table}[h]
\centering
\small
\begin{tabular}{lrrr}
\toprule
Step & OPSD & \shortstack[r]{\method{}\\w/o Residual} & \method{} \\
\midrule
20 & 56.5\ensuremath{\pm}0.2 & 66.3\ensuremath{\pm}0.4 & 51.7\ensuremath{\pm}0.5 \\
40 & 54.5\ensuremath{\pm}0.3 & 51.0\ensuremath{\pm}0.4 & 30.9\ensuremath{\pm}0.2 \\
60 & 55.8\ensuremath{\pm}0.7 & 55.3\ensuremath{\pm}0.7 & 46.9\ensuremath{\pm}2.3 \\
80 & 56.9\ensuremath{\pm}0.6 & 61.2\ensuremath{\pm}9.0 & 51.2\ensuremath{\pm}5.9 \\
100 & 56.9\ensuremath{\pm}0.1 & 61.3\ensuremath{\pm}2.6 & 60.8\ensuremath{\pm}9.3 \\
\midrule
Avg. & 56.1 & 59.0 & 48.3 \\
\bottomrule
\end{tabular}
\caption{Fraction of positive forward-KL mass removed by the student cap (\%). Checkpoint entries are means $\pm$ sample standard deviations across three training seeds. Avg. is the unweighted mean of the five checkpoint means.}
\label{tab:role-clipped-mass}
\end{table}

\section{Teacher--student style alignment}

\label{app:style-alignment}

\subsection{Experimental setup}

We tested whether teacher adaptation reduced teacher--student differences in response formatting. We used the fixed diagnostic set of 500 problems and the step-100 checkpoints of Qwen3-1.7B, Qwen3-4B, and Qwen3-8B. The \method{} checkpoints used residual weights $\beta=0.8$, $0.7$, and $0.5$. The student received only the problem, and the teacher received the problem and verified reference. The teacher and student used the same generation seed and decoding settings: temperature 1.0, top-$p$ 0.95, top-$k$ 20, a completion limit of 8,192 tokens, and a total limit of 16,384 tokens. At each model scale and training seed, we retained only problems for which all four responses in the OPSD and \method{} teacher--student pairs finished without truncation.

We measured formatting distance using nine features: newlines, paragraphs, Markdown headings, horizontal rules, bullets, numbered list items, bold spans, inline mathematics, and display mathematics. We normalized each feature count per 100 completion tokens and standardized it using a single comparison set that combines completions from all three model scales. The metric excludes answer correctness, content words, and the presence of a \texttt{\textbackslash boxed} answer. For problem $i$, the teacher--student formatting distance is

\begin{equation}
d_i^{\mathrm{fmt}}(T,S)=\frac{1}{9}\sum_{j=1}^{9}
\left|\frac{f_j(T_i)-f_j(S_i)}{s_j}\right|,
\label{eq:style-distance}
\end{equation}

where $f_j$ is the normalized count of feature $j$ and $s_j$ is its global sample standard deviation. The mean absolute difference weights the nine standardized features equally and is less sensitive to large individual discrepancies than a squared distance. Lower values indicate closer agreement in formatting. We averaged over retained problems within each training seed and report the mean and sample standard deviation across the three seeds.

\subsection{Step-100 quantitative results}

\begin{table}[h]
\centering
\small
\setlength{\tabcolsep}{3pt}
\begin{tabular}{lrrrr}
\toprule
Model & OPSD & \method{} & \shortstack[r]{Red.\\(\%)} & Retained \\
\midrule
1.7B & 0.823\ensuremath{\pm}0.022 & \textbf{0.467\ensuremath{\pm}0.049} & \textbf{43.2} & 449/423/463 \\
4B   & 0.810\ensuremath{\pm}0.006 & \textbf{0.266\ensuremath{\pm}0.069} & \textbf{67.2} & 486/483/470 \\
8B   & 0.830\ensuremath{\pm}0.007 & \textbf{0.390\ensuremath{\pm}0.039} & \textbf{53.0} & 469/479/457 \\
\bottomrule
\end{tabular}
\caption{Teacher--student formatting distance at step 100 for the three Qwen3 models. Lower values indicate closer agreement. Distances are means $\pm$ sample standard deviations across three training seeds. Red.\ is the relative reduction of \method{} from OPSD. Retained gives the numbers of problems for which all four compared responses were untruncated, in the listed training-seed order.}
\label{tab:style-alignment}
\end{table}

At step 100, \method{} had a lower formatting distance than OPSD at all three model scales (Table~\ref{tab:style-alignment}). The relative reductions were 43.2\% at 1.7B, 67.2\% at 4B, and 53.0\% at 8B. The largest reduction occurred at 4B, where the distance fell from 0.810 to 0.266. \method{} also had a lower distance in each of the nine model-scale and training-seed comparisons. These results show closer agreement in response layout and mathematical presentation between the adapted teacher and its contemporaneous student than between the fixed OPSD teacher and its student.

\subsection{Qualitative examples}

We show one example per model scale, selected from problems for which all four responses were untruncated and teacher--student formatting distance decreased. The full responses are reproduced below, with boldface marking portions relevant to the comparison.

\paragraph{Qwen3-1.7B (\texttt{openthoughts\_01993}. \textbf{distance: 0.808 to 0.229}).}
\textit{Problem:} What is the solution of the equation $24\div(3\div2)=(24\div3)\div m$?

\noindent\textbf{OPSD student output.}
\begin{quote}\small
We are given the equation:
\[
24 \div (3 \div 2) = (24 \div 3) \div m.
\]
We need to solve for $m$.

\textbf{Step 1: Simplify the left side of the equation}

\[
24 \div (3 \div 2).
\]
Recall that dividing by a fraction is the same as multiplying by its reciprocal:
\[
24 \div (3 \div 2) = 24 \times \frac{2}{3}.
\]
Now compute:
\[
24 \times \frac{2}{3} = \frac{24 \times 2}{3} = \frac{48}{3} = 16.
\]
So the left side simplifies to:
\[
16.
\]

\textbf{Step 2: Simplify the right side of the equation}

\[
(24 \div 3) \div m.
\]
Compute $24\div3$:
\[
24\div3=8.
\]
So the right side becomes:
\[
8\div m.
\]

\textbf{Step 3: Set the two sides equal}
\[
16=\frac{8}{m}.
\]

\textbf{Step 4: Solve for $m$}

Multiply both sides by $m$:
\[
16m=8.
\]
Now divide both sides by 16:
\[
m=\frac{8}{16}=\frac{1}{2}.
\]

\textbf{Final Answer:}
\[
\boxed{\frac{1}{2}}.
\]
\end{quote}

\noindent\textbf{OPSD teacher output.}
\begin{quote}\small
Let's solve the equation step by step, \textbf{using clear, step by step reasoning and avoiding copying the reference solution, but reproducing the same result through independent logical deduction.}

\textbf{Step 1: Understand the Equation}

We are given the equation:
\[
24 \div (3 \div 2) = (24 \div 3) \div m.
\]
Our goal is to find the value of $m$ that satisfies this equation.

\textbf{Step 2: Simplify the Left Hand Side (LHS)}

We begin with:
\[
24\div(3\div2).
\]
This is a division of a number by a fraction, which is equivalent to multiplying by the reciprocal of the fraction.
\[
24 \div \left(\frac{3}{2}\right)
=24\times\left(\frac{2}{3}\right).
\]
Now compute the multiplication:
\[
24\times\frac{2}{3}=\frac{24\times2}{3}=\frac{48}{3}=16.
\]
So, the LHS simplifies to:
\[
16.
\]

\textbf{Step 3: Simplify the Right Hand Side (RHS)}

We are now given:
\[
(24\div3)\div m.
\]
First, compute $24\div3$:
\[
24\div3=8.
\]
Now substitute back into the equation:
\[
8\div m=16.
\]
This is the simplified form of the RHS.

\textbf{Step 4: Solve for $m$}

We now have:
\[
8\div m=16.
\]
This is equivalent to:
\[
\frac{8}{m}=16.
\]
To solve for $m$, multiply both sides by $m$:
\[
8=16m.
\]
Then divide both sides by 16:
\[
m=\frac{8}{16}=\frac{1}{2}.
\]

\textbf{Final Answer}
\[
\boxed{\frac{1}{2}}.
\]
\end{quote}

\noindent\textbf{\method{} student output.}
\begin{quote}\small
\textbf{Let me solve the equation step by step:}

The equation is:
\[
24\div(3\div2)=(24\div3)\div m.
\]
First, let's simplify the left side:
\[
24\div(3\div2).
\]
Dividing by a fraction is the same as multiplying by the reciprocal:
\[
24\div(3\div2)=24\times(2\div3).
\]
Now compute $24\times(2\div3)$:
\[
24\times\frac{2}{3}=\frac{24\times2}{3}=\frac{48}{3}=16.
\]
So the left side is $16$.

Now the equation is:
\[
16=(24\div3)\div m.
\]
Compute $24\div3$:
\[
24\div3=8.
\]
So the equation is:
\[
16=8\div m.
\]
Now solve for $m$:
\[
16=\frac{8}{m}.
\]
Multiply both sides by $m$:
\[
16m=8.
\]
Divide both sides by 16:
\[
m=\frac{8}{16}=\frac{1}{2}.
\]
So the solution is $\boxed{\frac{1}{2}}$.
\end{quote}

\noindent\textbf{\method{} teacher output.}
\begin{quote}\small
\textbf{Let me solve the equation step by step.}

The equation is:
\[
24\div(3\div2)=(24\div3)\div m.
\]
First, let's simplify the left-hand side.

Left-hand side: $24\div(3\div2)$.

Dividing by a fraction is the same as multiplying by its reciprocal.

So:
\[
24\div(3\div2)=24\div\left(\frac{3}{2}\right).
\]
Dividing by a fraction is the same as multiplying by its reciprocal:
\[
24\div\left(\frac{3}{2}\right)=24\times\left(\frac{2}{3}\right).
\]
Now multiply:
\[
24\times\frac{2}{3}=\frac{24\times2}{3}=\frac{48}{3}=16.
\]
So left-hand side is 16.

Now simplify the right-hand side:

Right-hand side: $(24\div3)\div m$.

First, simplify $24\div3$:
\[
24\div3=8.
\]
So the right-hand side becomes:
\[
8\div m.
\]
So the equation is:
\[
16=8\div m.
\]
Now solve for $m$.
\[
8\div m=16.
\]
Dividing by $m$ is the same as multiplying by $1/m$.

So:
\[
8\div m=16.
\]
Multiply both sides by $m$:
\[
8=16m.
\]
Now solve for $m$:
\[
m=\frac{8}{16}.
\]
Simplify:
\[
m=\frac{1}{2}.
\]
So the solution is:
\[
\boxed{\frac{1}{2}}.
\]
\end{quote}

\noindent\textbf{Style observation.} The OPSD teacher introduces a teacher only meta instruction and a more elaborate heading structure. In contrast, the two \method{} responses share the same opening, paragraph rhythm, and predominantly inline presentation of intermediate equations.

\paragraph{Qwen3-4B (\texttt{openthoughts\_22848}. \textbf{distance: 0.971 to 0.185}).}
\textit{Problem:} Compute $9\cdot\frac{1}{13}\cdot26$.

\noindent\textbf{OPSD student output.}
\begin{quote}\small
Let's solve the expression step by step:

We start with the expression:
\[
9\cdot\frac{1}{13}\cdot26.
\]
First, multiply $\frac{1}{13}$ by 26:
\[
\frac{1}{13}\cdot26.
\]
Since 26 is divisible by 13, we can simplify this fraction:
\[
\frac{26}{13}=2.
\]
Now substitute this back into the original expression:
\[
9\cdot2.
\]
Now multiply 9 by 2:
\[
9\cdot2=18.
\]
So, the final answer is:
\[
\boxed{18}.
\]
\end{quote}

\noindent\textbf{OPSD teacher output.}
\begin{quote}\small
We are asked to compute the expression:
\[
9\cdot\frac{1}{13}\cdot26.
\]

\textbf{Step 1: Understand the structure of the expression}

We have a whole number (9), a fraction $\frac{1}{13}$, and another whole number (26). Since multiplication is commutative and associative, we can rearrange the terms in any order. This helps us group numbers that are easier to compute.

So, we can rearrange the expression as:
\[
9\cdot\left(\frac{1}{13}\cdot26\right).
\]

\textbf{Step 2: Simplify the fraction and whole number multiplication}

Let's compute the part inside the parentheses first:
\[
\frac{1}{13}\cdot26.
\]
This is the same as:
\[
\frac{26}{13}.
\]
Now simplify the fraction:
\[
\frac{26}{13}=2.
\]
So, the expression becomes:
\[
9\cdot2.
\]

\textbf{Step 3: Multiply the remaining numbers}

Now we just need to multiply 9 by 2:
\[
9\cdot2=18.
\]

\textbf{Final Answer:}
\[
\boxed{18}.
\]
\end{quote}

\noindent\textbf{\method{} student output.}
\begin{quote}\small
\textbf{Let's solve this step by step.}

First write down the problem:
\[
9\cdot(1/13)\cdot26.
\]
\textbf{Multiply $1/13$ by 26 first.}
\[
1/13\cdot26.
\]
26 divided by 13 is 2. So:
\[
1/13\cdot26=2.
\]
Now multiply by 9.
\[
9\cdot2=18.
\]
So the answer is 18.
\end{quote}

\noindent\textbf{\method{} teacher output.}
\begin{quote}\small
\textbf{Let's solve the problem step by step.}

The problem is: $9*(1/13)*26$.

\textbf{First, let's multiply $1/13$ and 26.}
\[
1/13*26=(1*26)/13=26/13.
\]
26 divided by 13 is 2.

So now we have:
\[
9*2.
\]
\[
9*2=18.
\]
The answer is 18.
\end{quote}

\noindent\textbf{Style observation.} The OPSD teacher expands the short calculation into a substantially more elaborate numbered template. The \method{} teacher instead matches the student's concise, unnumbered sequence and parallel wording.

\paragraph{Qwen3-8B (\texttt{openthoughts\_21633}. \textbf{distance: 0.986 to 0.417}).}
\textit{Problem:} What is the sum of $1+2+4+8+16+\cdots+1024$?

\noindent\textbf{OPSD student output.}
\begin{quote}\small
To find the sum of the series $1+2+4+8+16+\cdots+1024$, we can recognize that this is a geometric series. A geometric series is a series where each term is a constant multiple of the previous term.

\textbf{Step 1: Identify the first term and the common ratio}

The first term $a$ is $1$. The common ratio $r$ is $2$ (each term is twice the previous term).

\textbf{Step 2: Determine the number of terms}

The series is $1+2+4+8+16+\cdots+1024$. We need to find how many terms there are.

The $n$-th term of a geometric series is given by:
\[
a_n=a\cdot r^{n-1}.
\]
We know the last term $a_n=1024$, so we can solve for $n$:
\[
1024=1\cdot2^{n-1}.
\]
\[
1024=2^{n-1}.
\]
We know that $1024=2^{10}$, so:
\[
2^{10}=2^{n-1}.
\]
\[
10=n-1.
\]
\[
n=11.
\]
So, there are 11 terms in the series.

\textbf{Step 3: Use the formula for the sum of a geometric series}

The sum $S_n$ of the first $n$ terms of a geometric series is given by:
\[
S_n=a\cdot\frac{r^n-1}{r-1}.
\]
Substitute $a=1$, $r=2$, and $n=11$:
\[
S_{11}=1\cdot\frac{2^{11}-1}{2-1}.
\]
\[
S_{11}=\frac{2^{11}-1}{1}.
\]
\[
S_{11}=2^{11}-1.
\]

\textbf{Step 4: Calculate the result}
\[
2^{11}=2048.
\]
\[
S_{11}=2048-1=2047.
\]

\textbf{Final Answer:}
\[
\boxed{2047}.
\]
\end{quote}

\noindent\textbf{OPSD teacher output.}
\begin{quote}\small
Let's break down the problem step by step and think through the logic to arrive at the answer.

\textbf{Step 1: Understand the Series}

We are given the series:
\[
1+2+4+8+16+\cdots+1024.
\]
This is clearly a geometric series. In a geometric series, each term is obtained by multiplying the previous term by a constant, called the common ratio.

Let's identify the components:

First term, $a=1$; common ratio, $r=2$ (since each term is double the previous one); and the last term is 1024. Let's verify this is part of the pattern:
\[
1,2,4,8,16,32,64,128,256,512,1024.
\]
So the series is:
\[
2^0,2^1,2^2,\dots,2^{10}.
\]
That means the number of terms is $n=11$.

\textbf{Step 2: Use the Formula for the Sum of a Finite Geometric Series}

The formula for the sum of the first $n$ terms of a geometric series is:
\[
S_n=\frac{a(1-r^n)}{1-r}.
\]
We have already found $a=1$, $r=2$, and $n=11$.

Plug these values into the formula:
\[
S_{11}=\frac{1(1-2^{11})}{1-2}.
\]

\textbf{Step 3: Simplify the Expression}

First, compute the numerator and denominator:

Numerator: $1-2^{11}=1-2048=-2047$.

Denominator: $1-2=-1$.

Now divide:
\[
S_{11}=\frac{-2047}{-1}=2047.
\]

\textbf{Step 4: Final Answer}

So the sum of the series:
\[
1+2+4+8+16+\cdots+1024=\boxed{2047}.
\]

\textbf{Conclusion}

The series is a geometric series with first term $a=1$, common ratio $r=2$, and 11 terms. Using the geometric series sum formula, we find the total sum to be 2047.
\end{quote}

\noindent\textbf{\method{} student output.}
\begin{quote}\small
\textbf{The sum of $1+2+4+8+16+\cdots+1024$ is a geometric series.}

A geometric series is a series where each term is the previous term multiplied by a common ratio.

In this series:
\[
1,2,4,8,16,\ldots,1024.
\]
The first term is 1.

The common ratio is 2 (each term is the previous term multiplied by 2).

The last term is 1024.

The formula for the sum of a geometric series is:
\[
\begin{aligned}
\text{Sum}
&=(\text{first term}*(\text{common ratio}^n-1))\\
&\quad/(\text{common ratio}-1).
\end{aligned}
\]
Where $n$ is the number of terms.

So, we need to find $n$.

The $n$th term of a geometric series is:
\[
\text{term}_n=\text{first term}*(\text{common ratio})^{n-1}.
\]
In this case, $\text{term}_n$ is 1024.

So:
\[
1024=1*2^{n-1}.
\]
So $2^{n-1}=1024$.

We know that $2^{10}=1024$.

So:
\[
2^{n-1}=2^{10}.
\]
So $n-1=10$.

So $n=11$.

So there are 11 terms.

Now, calculate the sum:
\[
\text{Sum}=(1*(2^{11}-1))/(2-1).
\]
\[
=(2^{11}-1)/1.
\]
\[
=2048-1.
\]
\[
=2047.
\]
So the sum is 2047.
\end{quote}

\noindent\textbf{\method{} teacher output.}
\begin{quote}\small
\textbf{Let's see. The problem is to calculate the sum of $1+2+4+8+16+\cdots+1024$.}

Hmm, I notice that each term is double the previous one. So this is a geometric series.

The formula for the sum of a geometric series is:
\[
S=a*(r^n-1)/(r-1).
\]
Where $a$ is the first term, $r$ is the common ratio, and $n$ is the number of terms.

In this case, the first term $a$ is 1. The common ratio $r$ is 2 (each term is double the previous one). The last term is 1024.

So we need to find $n$, the number of terms.

The $n$th term of a geometric series is $a*r^{n-1}$.

So in this case, the last term is $1*2^{n-1}=1024$.

So $2^{n-1}=1024$.

What's 1024? It's $2^{10}$.

So $2^{n-1}=2^{10}$.

So $n-1=10$.

$n=11$.

So there are 11 terms.

So the sum is:
\[
S=1*(2^{11}-1)/(2-1).
\]
\[
=(2048-1)/1.
\]
\[
=2047.
\]
So the sum is 2047.
\end{quote}

\noindent\textbf{Style observation.} The OPSD pair uses different heading inventories and the teacher adds a separate conclusion. The \method{} pair instead uses the same prose first organization, short paragraphs, and minimally formatted sequence from identifying the ratio through computing the final sum.

\end{document}